\documentclass{article}
\usepackage{iclr2027_conference,times}
\iclrfinalcopy 

\usepackage[utf8]{inputenc} 
\usepackage[T1]{fontenc}    
\usepackage{hyperref}       
\usepackage{url}            
\usepackage{booktabs}       
\usepackage{amsfonts}       
\usepackage{nicefrac}       
\usepackage{microtype}      
\usepackage{xcolor}         
\usepackage{booktabs}
\usepackage{multirow}
\usepackage{colortbl}
\usepackage{graphicx}
\usepackage{amsmath}
\usepackage{subcaption}
\usepackage{fvextra}
\title{From Priors to Perception: Grounding Video-LLMs in Physical Reality}

\author{%
  \normalfont
  \begin{tabular}[t]{@{}p{0.45\textwidth}@{\hspace{0.06\textwidth}}p{0.45\textwidth}@{}}
    \textbf{Zicheng Zhao} & \textbf{Chaofan Gan} \\
    Shanghai Jiao Tong University & Shanghai Jiao Tong University \\
    \texttt{zzcsjtu7@sjtu.edu.cn} & \texttt{ganchaofan@sjtu.edu.cn} \\[1.5em]
    \textbf{Shijie Li} & \textbf{Weiyao Lin} \\
    Shanghai Jiao Tong University & Shanghai Jiao Tong University \\
    \texttt{shijieli@sjtu.edu.cn} & \texttt{wylin@sjtu.edu.cn}
  \end{tabular}
}

\begin{document}
\maketitle
\lhead{} 
\begin{abstract}
Video Large Language Models (Video-LLMs) excel in general video understanding but often base physical judgments on event expectations rather than observations. We find that they not only rationalize physically impossible events, but also incorrectly report expected outcomes in physically plausible yet counter-intuitive scenarios, despite clear contradictory visual evidence. We provide the first unified account of these failures as Semantic Prior Dominance (SPD): semantic expectations override conflicting visual evidence. To diagnose and overcome these failures, we introduce PriorPair, a high-fidelity paired video benchmark covering both conflicts across eight physical categories. Its category-directed construction pairs prior-aligned events with semantically matched but prior-conflicting counterparts. We further propose Physics-Anchored Reasoner (PhyAR), a lightweight training framework. Its Visually Anchored Reasoning Chain grounds physical attribution and judgment in explicit visual observations, while Paired Data Binding strengthens learning of decisive event differences through matched-pair supervision. Standard LoRA fine-tuning with PhyAR substantially improves physical reasoning and reduces prior-driven judgment bias across different Video-LLM backbones without architectural modifications. These gains extend to an external physical benchmark while largely preserving general video understanding.
\par\smallskip
\noindent{\small\url{https://github.com/LiamZhao326/From-Priors-to-Perception}}
\end{abstract}

\section{Introduction}
\label{inst}
Video Large Language Models (Video-LLMs) excel at general video understanding \citep{lin2024video,maaz2024video}, yet struggle with fine-grained physical reasoning, failing to reliably recognize even object levitation and rigid-body penetration \citep{bai2025impossible,motamed2025travl}.
Beyond physical common sense, this reasoning requires integrating visual perception, event dynamics, and causal relations to judge what actually happens.
As AI advances toward world models and embodied AI, grounding judgments in visual facts rather than expectations is essential for reliable scene simulation and real-world interaction.
We therefore ask: when observations conflict with prior expectations, do Video-LLMs base physical judgments on visual evidence?

\begin{figure}[t]
    \centering
    \begin{subfigure}{\linewidth}
        \centering
        \includegraphics[width=\linewidth]{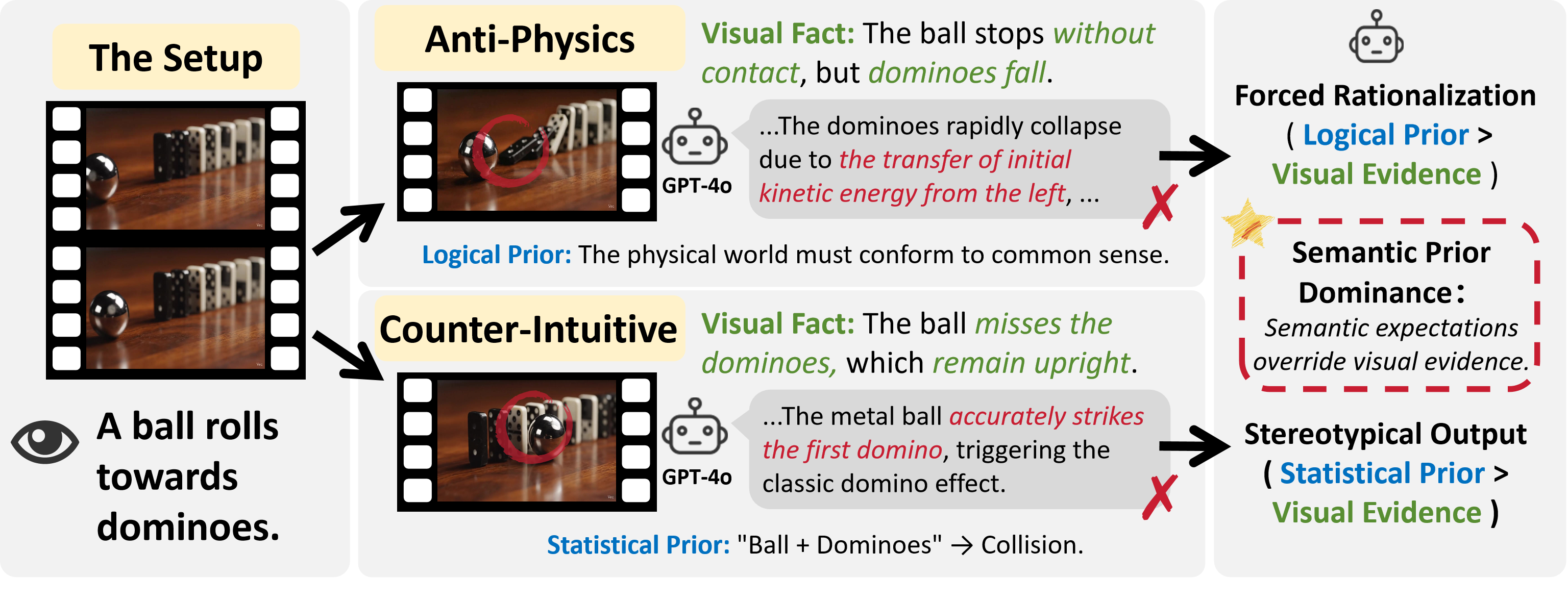}
        \vspace{-7mm} 
        \caption{}    
        \label{fig:intro_a}
    \end{subfigure}
    
    \vspace{-1mm} 
    
    \begin{subfigure}{0.48\linewidth}
        \centering
        \includegraphics[width=\linewidth]{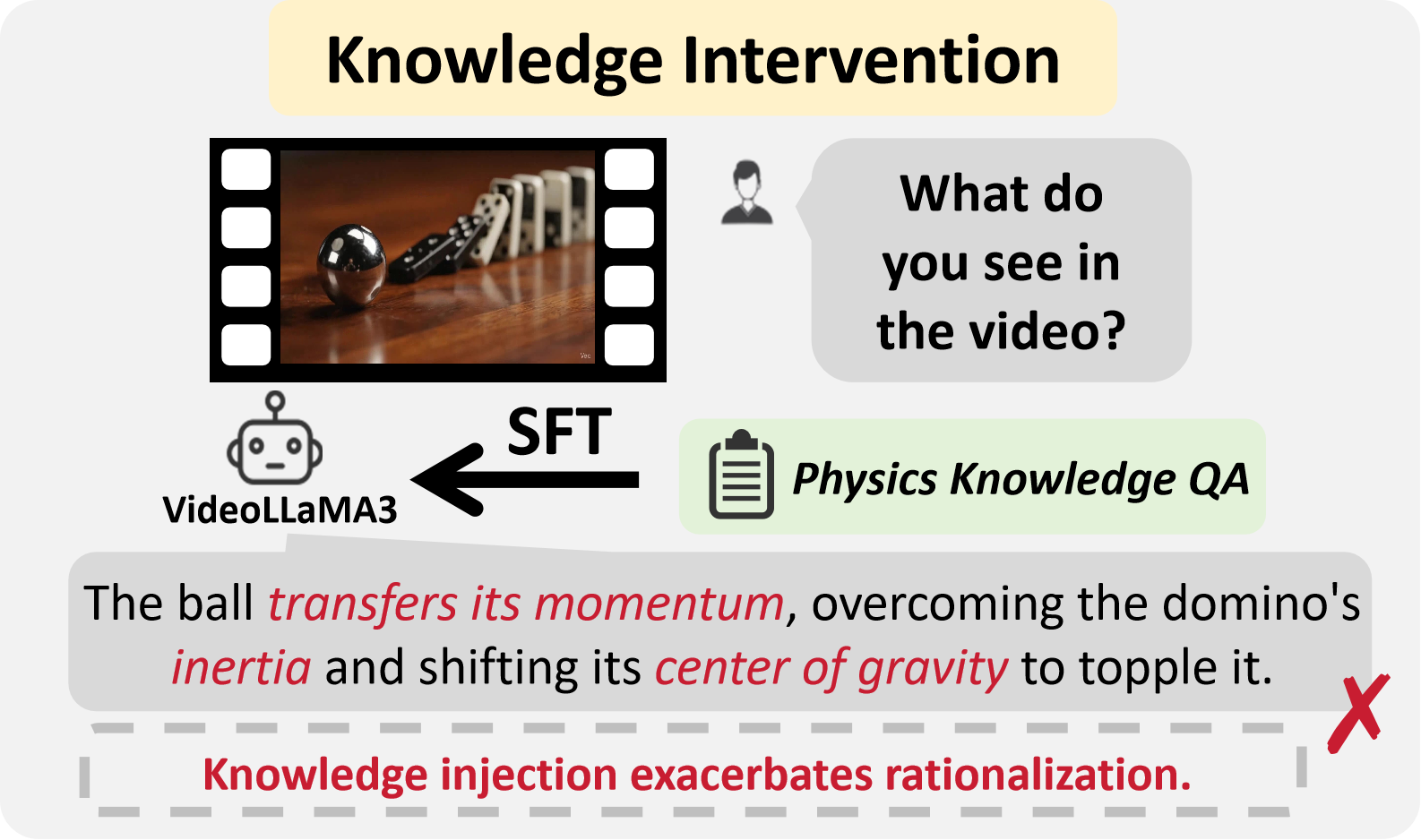}
        \vspace{-5mm} 
        \caption{}    
        \label{fig:intro_b}
    \end{subfigure}
    \hfill
    \begin{subfigure}{0.48\linewidth}
        \centering
        \includegraphics[width=\linewidth]{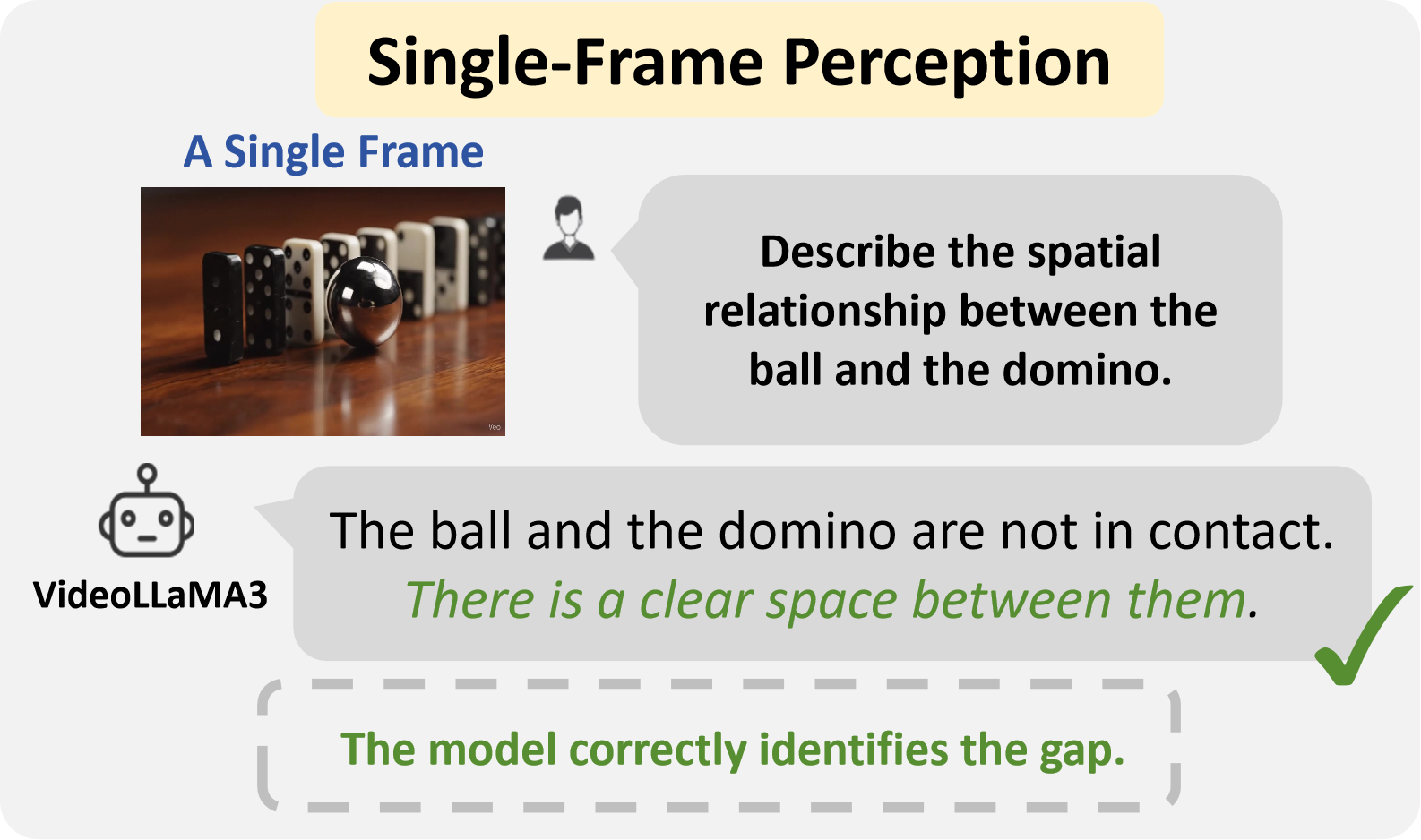}
        \vspace{-5mm} 
        \caption{}    
        \label{fig:intro_c}
    \end{subfigure}
    
    \vspace{-2mm} 
    
    \caption{\textbf{Illustration of Semantic Prior Dominance.} \textbf{(a)} GPT-4o~\citep{openai2024gpt4o} rationalizes anomalies in \textit{anti-physics} scenarios and describes expected events despite conflicting visual evidence in \textit{counter-intuitive} scenarios. Two pilots with VideoLLaMA3~\citep{zhang2025videollama3} show that \textbf{(b)} physics-knowledge QA fine-tuning exacerbates erroneous rationalization, whereas \textbf{(c)} a single-frame spatial query recovers the gap between the ball and dominoes.}
    \label{fig:teaser}
    \vspace{-19pt}
\end{figure}

A ball--domino interaction illustrates these failures (Fig.~\ref{fig:intro_a}). In an \textit{anti-physics} scenario, dominoes fall without ball contact, yet GPT-4o \citep{openai2024gpt4o} fabricates kinetic-energy transfer to rationalize the anomaly. This \textit{forced rationalization} reflects a \textit{logical-consistency prior}: ``the world must conform to common sense.'' Models fail even in physically plausible yet \textit{counter-intuitive} real-world scenarios: when the ball misses the dominoes, leaving them standing, the model ignores the visible gap and follows the \textit{statistical event prior} that balls typically topple dominoes, producing \textit{what-it-expects-is-what-it-sees} stereotypical output. We further examine these failures through two pilots with VideoLLaMA3 \citep{zhang2025videollama3}. Physics-knowledge QA fine-tuning intensifies rationalization: the model uses correct physical concepts to fabricate a more elaborate account of a nonexistent collision (Fig.~\ref{fig:intro_b}), yet it correctly identifies the ball--domino gap in a single-frame query (Fig.~\ref{fig:intro_c}).

These observations reveal a crucial mismatch: relevant knowledge and decisive visual evidence are accessible, yet the model privileges semantic expectations over visual facts. We call this phenomenon \textbf{Semantic Prior Dominance (SPD)}, which provides a unified behavioral account of models' forced rationalization in anti-physics scenarios and stereotypical descriptions in counter-intuitive scenarios.

Existing physical reasoning research still faces key gaps. First, physical implausibility detection \citep{motamed2025travl} and hallucination of expected events \citep{lee2025noah} are studied separately. Second, physical anomaly datasets often collect stochastic failures produced by video generators \citep{bai2025impossible,motamed2025travl}. Generation stochasticity limits control over failure types and reliable, targeted coverage of predefined physical categories. Artifacts such as texture flickering or unintended deformation can also offer shortcuts around decisive physical relations. Third, interventions often depend on specific physical representations (such as motion trajectories), model architectures, or individual tasks \citep{motamed2025travl,zhan2025phyvllm,li2025videohallu}, limiting the range of physical scenarios they can address. A unified paradigm is still missing that combines category-directed, high-fidelity data with architecture-compatible training to address both prior conflicts and cover the diverse physical phenomena they involve.

We introduce \textbf{PriorPair}, a paired video benchmark for physical reasoning when semantic priors conflict with visual evidence. Paired videos share similar scene semantics but depict prior-aligned and prior-conflicting events, respectively. Guided by a top-down taxonomy of eight physical categories, we construct high-fidelity pairs for each category and provide complete \textbf{Observation--Attribution--Verdict (O--A--V)} annotations for every video. It comprises two adversarial streams addressing the two prior conflicts: \textit{Anti-physics} follows the principle of \textit{Visually Plausible, Physically Invalid}, injecting explicit physical violations into high-quality scenes; \textit{Counter-intuitive} constructs \textit{Physically Plausible, Statistically Unexpected} events, requiring models to report observed events rather than impose familiar event scripts. By reducing incidental visual variation, PriorPair further focuses evaluation and training on decisive visual differences. Models cannot correctly judge both videos in a pair from shared semantic expectations alone. Across models and physical categories, accuracy on PriorPair is substantially higher for prior-aligned videos than for their prior-conflicting counterparts, showing that SPD is a systematic behavioral phenomenon. Building on PriorPair, we propose \textbf{PhyAR} (\textbf{Phy}sics-\textbf{A}nchored \textbf{R}easoner), a lightweight, architecture-compatible training framework. Its \textbf{Visually Anchored Reasoning Chain (VARC)} uses these annotations for structured supervision, guiding models from observed facts through physical interpretation to judgment. \textbf{Paired Data Binding} binds counterparts for joint supervision within the same parameter update. Together, structured reasoning and paired supervision strengthen judgments grounded in observed events.

Our main contributions are as follows:
\begin{itemize}
    \item We reveal \textbf{Semantic Prior Dominance}, unifying for the first time models' forced rationalization in Anti-physics scenarios and stereotypical reporting of expected events in Counter-intuitive scenarios.
    \item We construct \textbf{PriorPair}, the first paired video benchmark and training set to jointly cover both prior conflicts, with high-fidelity matched video pairs spanning eight physical categories for strict paired evaluation and visually grounded learning.
    \item We propose \textbf{PhyAR}, combining VARC and Paired Data Binding in a lightweight, architecture-compatible LoRA fine-tuning framework \citep{hu2021lora} that substantially improves physical reasoning on two different backbones and transfers to an external physical benchmark while largely preserving general video understanding.
\end{itemize}

\section{Related Work}
\label{sec:rel}
\paragraph*{Video Construction.}
Current video generators provide diverse synthetic scenes \citep{blattmann2023stable,brooks2024videogeneration,team2025kling,veo2025}. Impossible Videos \citep{bai2025impossible}, TRAVL \citep{motamed2025travl}, and WorldModelBench \citep{li2025worldmodelbench} use generated videos to study physical consistency and failures. However, stochastic video generation limits control over natural or prompt-induced failure types and makes diverse physical scenarios difficult to cover adequately. Generator artifacts can also provide shortcuts that bypass decisive physical relations. Guided by a physical taxonomy, PriorPair uses targeted video editing or constrained generation to construct prior-conflicting counterparts around specified physical relations or event outcomes, combining systematic category coverage with high-fidelity scenes. Experts verify visual quality, the intended phenomenon, counterpart alignment, and annotation logic. Paired construction and expert verification together reduce incidental visual variation and opportunities for artifact shortcuts.

\paragraph*{Physical Reasoning Benchmarks.}
Video-LLMs have advanced multiple aspects of video understanding \citep{lin2024video,maaz2024video,chen2024longvila,wang2025internvideo25,zhao2026cogstream}, while physical reasoning and video hallucination benchmarks examine complementary aspects of their reliability. The former assess physical validity, event dynamics, and anomaly detection \citep{riochet2018intphys,yi2020clevrer,chow2025physbench,hong2025motionbench,bai2025impossible,motamed2025travl,li2025videohallu}; the latter evaluate hallucinations and omissions in video question answering and captioning \citep{wang2024videohallucer,rawal2025argus} and examine the role of familiar event patterns and expectations of narrative coherence in these errors \citep{zhang2024eventhallusion,lee2025noah}. Yet forced rationalization of impossible events and hallucination of expected events remain largely studied and evaluated separately. We reveal Semantic Prior Dominance (SPD), which provides the first unified behavioral account of forced rationalization of impossible events and stereotypical reporting of expected outcomes in physically plausible yet counter-intuitive scenarios, and introduce PriorPair to jointly evaluate Anti-physics and Counter-intuitive prior conflicts within a unified paired physical reasoning benchmark. It requires correct judgments of both prior-aligned and prior-conflicting counterparts to test whether models remain faithful to the video's visual evidence when priors conflict with observations.

\paragraph*{Reasoning Interventions.}
Current interventions for physical reasoning span architectural design, inference-time decoding, and post-training. Architectural approaches use trajectory-aware representations \citep{patrick2021motionformer,motamed2025travl} or introduce Neural ODEs \citep{zhan2025phyvllm,chen2018neuralode} to enhance dynamics modeling, but these motion-focused architectural changes are insufficient to address diverse physical failures. Inference-time contrastive decoding improves temporal faithfulness but adds inference overhead \citep{wu2025season}. Post-training includes robust instruction tuning \citep{liu2024mitigating}, preference optimization \citep{rafailov2023direct}, and reinforcement learning \citep{shao2024deepseekmath}. Task-specific reward design constrains the extension of video reinforcement learning to new tasks \citep{li2025videohallu,feng2025videor1,li2025videochatr1}. PhyAR combines PriorPair's systematic coverage of eight physical categories across both streams with VARC and Paired Data Binding for visually grounded supervision. Standard LoRA fine-tuning \citep{hu2021lora} improves physical reasoning without modifying backbone architectures, introducing category-specific modules, or designing reward functions. PhyAR achieves substantial and consistent gains across backbones and transfers to IntPhys2 \citep{bordes2025intphys2}.

\section{PriorPair: Diagnosing Prior-Conflicting Physical Reasoning}
\label{sec:priorpair}
\vspace{-4pt}
\subsection{Paired Task Design and Taxonomy}
\label{sec:taxonomy}
\vspace{-4pt}

\begin{figure*}[t]
  \centering
  \includegraphics[width=\linewidth]{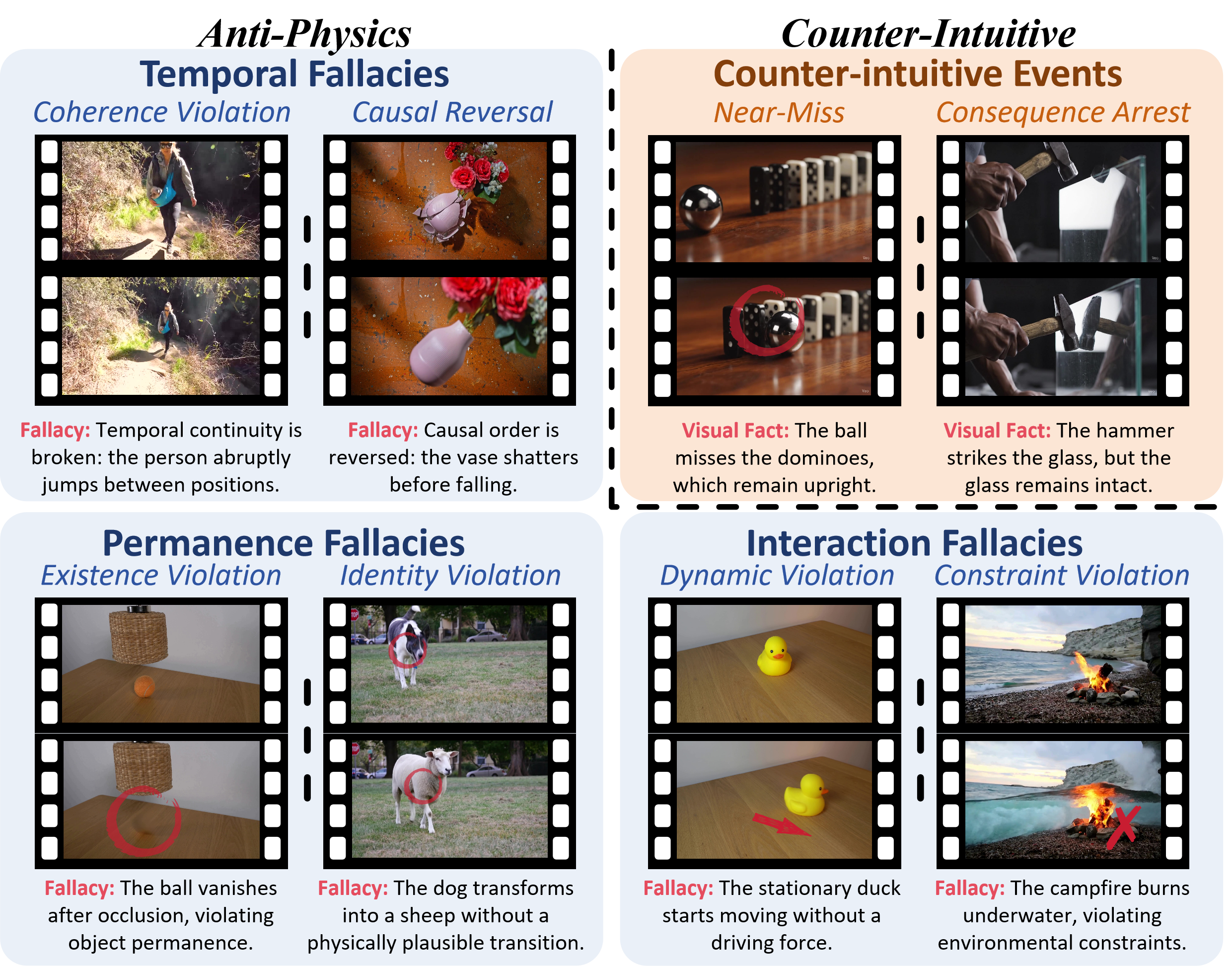}
  \vspace{-15pt}
  \caption{\textbf{Taxonomy of PriorPair: two streams, four groups, eight categories.} For each category, frames from a representative negative counterpart are arranged chronologically from top to bottom. Anti-physics negatives violate physical laws, whereas Counter-intuitive negatives remain physically plausible but lack the expected target event. Red annotations highlight decisive visual evidence.}
  \label{fig:taxonomy}
  \vspace{-12pt}
\end{figure*}

\textbf{PriorPair} is a dual-stream paired video benchmark for physical reasoning when visual evidence conflicts with semantic priors. Each pair comprises a \textit{positive anchor} following the expected course of events and a matched \textit{negative counterpart} defying this expectation. In the \textbf{Anti-physics} stream, the positive and negative videos are physically valid and invalid, respectively, and the task is \textit{physical-validity verification}. In the \textbf{Counter-intuitive} stream, both videos are physically valid, but the expected target event occurs only in the positive video; the task is \textit{target-event occurrence verification}. Paired videos share similar scene semantics but differ in core physical relations or event outcomes, requiring different verdicts. The same semantic expectation cannot yield correct judgments for both videos. By aligning scene semantics and reducing incidental visual variation, PriorPair focuses evaluation and training on the ability to reason and make judgments grounded in actual observations.

PriorPair adopts a \textbf{top-down taxonomy} that specifies target phenomena before construction and guides source selection, editing, and annotation to systematically cover physical phenomena rather than collect unstructured generative failures. This taxonomy draws on developmental psychology's Core Knowledge theories \citep{spelke1990principles,baillargeon2002physical} and intuitive-physics benchmarks IntPhys and CLEVRER \citep{riochet2018intphys,yi2020clevrer}. The Anti-physics stream comprises \textit{Temporal Fallacies}, \textit{Permanence Fallacies}, and \textit{Interaction Fallacies}. The Counter-intuitive stream forms the fourth group: \textit{Counter-intuitive Events}. These four groups span eight categories (Fig.~\ref{fig:taxonomy}).

\noindent\textbf{Temporal Fallacies.} These categories violate temporal continuity and causal order. (1) \textit{Coherence Violation} disrupts local temporal continuity, e.g., a person abruptly jumping between positions in shuffled walking frames. (2) \textit{Causal Reversal} inverts the macroscopic cause-and-effect order, e.g., a vase shattering before falling.

\noindent\textbf{Permanence Fallacies.} These categories violate object persistence and property stability. (3) \textit{Existence Violation} makes objects or matter appear or disappear without plausible physical processes, e.g., a ball vanishing after occlusion. (4) \textit{Identity Violation} abruptly changes object identity, shape, quantity, or intrinsic properties without plausible physical transitions, e.g., a dog becoming a sheep.

\noindent\textbf{Interaction Fallacies.} These categories violate dynamics or physical interaction constraints. (5) \textit{Dynamic Violation} makes motion, force balance, or thermodynamic changes inconsistent with the governing physical conditions, e.g., a stationary toy duck starting to move without a driving force. (6) \textit{Constraint Violation} breaches spatial boundaries, material properties, conservation relations, or constraints linking an interaction to its consequences, e.g., a campfire burning underwater.

\noindent\textbf{Counter-intuitive Events.} These categories remain physically plausible while the expected event fails to occur. (7) \textit{Near-Miss} places an object's trajectory within an interaction zone, but a spatial gap prevents the expected contact, e.g., a ball passing beside dominoes without contact and leaving them upright. (8) \textit{Consequence Arrest} establishes the initiating action or contact, but insufficient force or other factors prevent the expected consequence, e.g., a hammer striking glass without breaking it. Detailed definitions and further examples for all eight categories are in Appendices~\ref{appendix:taxonomy_antiphysics} and~\ref{appendix:taxonomy_counterintuitive}.

\subsection{Construction and Quality Control}

\begin{figure}[t]
  \centering
  \includegraphics[width=\linewidth]{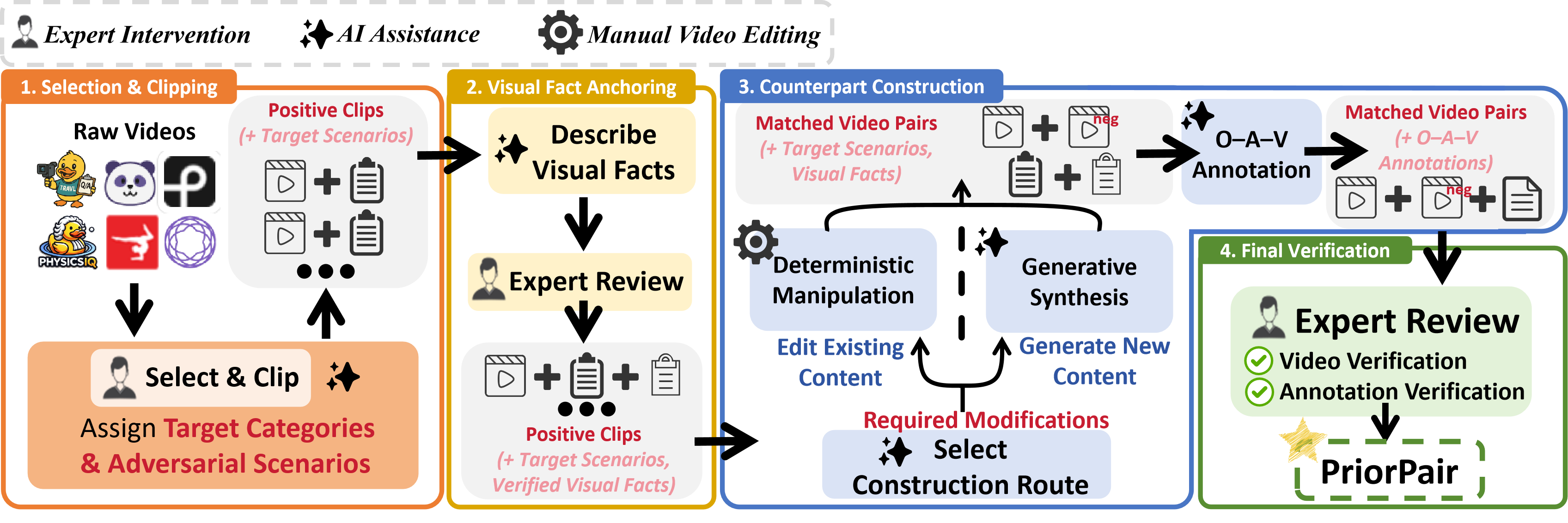} 
  \vspace{-14pt}
  \caption{\textbf{PriorPair construction through a four-stage human-in-the-loop pipeline.}}
  \label{fig:pipeline}
  \vspace{-14pt}
\end{figure}

PriorPair combines video generation with precise manual editing in a four-stage \textbf{Human-in-the-loop (HITL)} pipeline to construct diverse adversarial scenarios (Fig.~\ref{fig:pipeline}). Source videos are curated from Something-Something V2 \citep{goyal2017something}, PhysBench \citep{chow2025physbench}, Panda-70M \citep{chen2024panda}, ImplausiBench \citep{motamed2025travl}, ActivityNet \citep{caba2015activitynet}, Physics-IQ \citep{motamed2026generative}, and Pexels\footnote{\url{https://www.pexels.com/}}.

\noindent\textbf{Stage 1: Selection \& Clipping.} Experts select and trim high-quality videos into positive anchors depicting singular physical interactions, then assign target categories and adversarial scenarios with LLM assistance. \textbf{Stage 2: Visual Fact Anchoring.} Multimodal LLMs describe positive anchors' visual facts. Experts remove visual hallucinations to ensure \textit{purity} and exclude physical speculation, such as \textit{falling due to gravity}, to maintain \textit{objectivity}, yielding verified visual fact anchors. \textbf{Stage 3: Counterpart Construction.} Given the positive anchor, visual facts, and target scenario, the pipeline selects between two paths according to required modifications, producing a matched negative counterpart. \textit{Deterministic manipulation} rearranges or edits existing content through operations such as frame reordering and mask-based removal, performed by experts with SAM 2 \citep{ravi2024sam2} and ProPainter \citep{zhou2023propainter} assisting localized editing. \textit{Generative synthesis} uses carefully tailored, target-specific prompts to guide video models such as Kling \citep{team2025kling} in generating new visual content or altering physical properties, such as morphological changes or trajectory deviations. LLMs generate complete Observation--Attribution--Verdict annotations per video (Sec.~\ref{sec:varc}), with both videos in a Counter-intuitive pair sharing the same target event. \textbf{Stage 4: Final Verification.} Experts review video pairs and annotations for video quality, intended phenomena, and annotation logic. Construction details and prompts appear in Appendices~\ref{appendix:construction_quality} and~\ref{appendix:prompt_scenario}--\ref{appendix:prompt_annotation}.

\noindent\textbf{Dataset Statistics and Properties.} PriorPair contains \textbf{828 matched video pairs} (\textbf{660 training}, \textbf{168 test}), with videos averaging \textbf{4.0 seconds}. Category proportions: Coherence (10.14\%), Causal Reversal (10.14\%), Existence (5.19\%), Identity (5.80\%), Dynamic (29.59\%), Constraint (14.73\%), Near-Miss (13.29\%), and Consequence Arrest (11.11\%). Three experts independently reviewed each pair's video quality, target phenomena, correspondence, category assignments, O--A--V annotations, and unintended visual cues. PriorPair prioritizes high fidelity and expert verification; its reusable taxonomy, routing procedure, and construction templates support scaling under expert quality control.

\subsection{Evaluation Protocol}
\label{sec:evaluation_protocol}

For $N$ pairs, $c_i^+,c_i^-\in\{0,1\}$ indicate correct verdicts for positive and negative videos, respectively. \textbf{Point-wise Accuracy} measures single-video accuracy; our primary metric, \textbf{Pair-wise Consistency Accuracy (PCA)}, requires correct verdicts for both videos in a pair:
\begin{equation}
\mathrm{PointAcc}=\frac{100}{2N}\sum_{i=1}^{N}(c_i^++c_i^-),
\qquad
\mathrm{PCA}=\frac{100}{N}\sum_{i=1}^{N}c_i^+c_i^-.
\end{equation}
PCA prevents high accuracy on one side from masking failures on the other.

\textbf{Relative Prior Bias (RPB)} measures the positive--negative accuracy gap via Michelson contrast:
\begin{equation}
\mathrm{RPB}=100\times
\frac{|\mathrm{Acc}_{\mathrm{pos}}-\mathrm{Acc}_{\mathrm{neg}}|}
{\mathrm{Acc}_{\mathrm{pos}}+\mathrm{Acc}_{\mathrm{neg}}},
\end{equation}
where $\mathrm{Acc}_{\mathrm{pos}}$ and $\mathrm{Acc}_{\mathrm{neg}}$ are positive and negative point-wise accuracies, respectively. Lower RPB indicates more balanced performance; the respective accuracies reveal bias direction.

\textbf{Reasoning Alignment Score (RAS)} uses an LLM judge to rate visual observation and physical attribution quality on a 1--5 scale. See Appendix~\ref{appendix:ras_protocol} for the full evaluation protocol.

\section{PhyAR}
\label{method}

To address Semantic Prior Dominance, we propose \textbf{Phy}sics-\textbf{A}nchored \textbf{R}easoner (PhyAR), an architecture-compatible training framework combining \textbf{Visually Anchored Reasoning Chain (VARC)} and \textbf{Paired Data Binding}. VARC grounds physical reasoning in explicit visual observations, while Paired Data Binding jointly supervises matched videos from PriorPair.

\subsection{Visually Anchored Reasoning Chain}
\label{sec:varc}

When priors conflict with visual evidence, Video-LLMs may base judgments on priors activated by the question and scene rather than decisive visual facts. VARC organizes responses into Observation--Attribution--Verdict sequences, placing explicit visual observations before physical attribution and final judgment. \textbf{Observation} identifies entities, spatial relations, contact states, temporal order, and state changes relevant to the target event (e.g., the ball--domino gap). \textbf{Attribution} uses existing physical knowledge to interpret these facts in terms of dynamics, states, spatiotemporal and causal relations, and physical constraints, e.g., identifying the constraints violated by fire burning underwater rather than inventing unseen factors to rationalize it. \textbf{Verdict} determines from these observations and attribution whether the video is physically valid or the target event occurred.

\begin{table*}[t]
\centering
\small
\setlength{\tabcolsep}{3.9pt}
\renewcommand{\arraystretch}{0.95}
\caption{\textbf{PCA (\%) and RAS (1--5) on PriorPair.} Overall PCA is micro-averaged over all 168 test pairs. \textit{Anti-physics}: Coh. (Coherence), Cau. (Causal Reversal), Exi. (Existence), Ide. (Identity), Dyn. (Dynamic), and Cns. (Constraint). \textit{Counter-intuitive}: Near-Miss and C-Arrest (Consequence Arrest). Best and second-best results are \textbf{bold} and \underline{underlined}. Category-level RAS appears in Appendix~\ref{appendix:ras_results}.}
\label{tab:main_results}
\vspace{-4pt}
\fontsize{8}{9.5}\selectfont
\begin{tabular}{l cccccc cc cc}
\toprule
\multirow{2}{*}{\textbf{Method}} & \multicolumn{6}{c}{\textbf{Anti-physics}} & \multicolumn{2}{c}{\textbf{Counter-intuitive}} & \multicolumn{2}{c}{\textbf{Overall}} \\
\cmidrule(lr){2-7} \cmidrule(lr){8-9} \cmidrule(lr){10-11}
 & \textbf{Coh.} & \textbf{Cau.} & \textbf{Exi.} & \textbf{Ide.} & \textbf{Dyn.} & \textbf{Cns.} & \textbf{Near-Miss} & \textbf{C-Arrest} & \textbf{PCA}$\uparrow$ & \textbf{RAS}$\uparrow$ \\
\midrule
\multicolumn{11}{c}{\textit{Proprietary Models}} \\
\midrule

GPT-4o & 17.65 & 17.65 & 0.00 & \textbf{80.00} & 26.53 & \underline{32.00} & 31.82 & 52.63 & 30.95 & 2.440 \\
Gemini 2.5 Flash & 11.76 & 35.29 & 0.00 & 60.00 & \underline{38.78} & \textbf{36.00} & 22.73 & 42.11 & 32.74 & 2.330 \\
\midrule
\multicolumn{11}{c}{\textit{Open-Source Models}} \\
\midrule
Flash-VStream-7B & 5.88 & 5.88 & 11.11 & 20.00 & 8.16 & 12.00 & 0.00 & 10.53 & 8.33 & 1.250 \\
Video-ChatGPT-7B & 0.00 & 0.00 & 0.00 & 0.00 & 0.00 & 0.00 & 13.64 & 0.00 & 1.79 & 1.101 \\
Video-LLaVA-7B & 11.76 & 11.76 & 11.11 & 40.00 & 6.12 & 16.00 & 4.55 & 10.53 & 11.31 & 1.289 \\
MiniCPM-V 2.6 & 5.88 & 0.00 & 0.00 & 10.00 & 2.04 & 0.00 & 4.55 & 47.37 & 7.74 & 1.518 \\
VideoLLaMA2-7B & 0.00 & 0.00 & 0.00 & 0.00 & 0.00 & 0.00 & 27.27 & 26.32 & 6.55 & 1.399 \\
Qwen3-VL-8B-Instruct & 0.00 & 5.88 & 0.00 & \underline{70.00} & 16.33 & 20.00 & 31.82 & 47.37 & 22.02 & 1.982 \\
VideoLLaMA3-7B & 0.00 & 0.00 & 0.00 & 30.00 & 2.04 & 4.00 & 22.73 & \underline{68.42} & 13.69 & 1.768 \\
\rowcolor{gray!15}
\textbf{PhyAR (VideoLLaMA3-7B)} & \underline{41.18} & \underline{47.06} & \underline{22.22} & 40.00 & \underline{38.78} & 28.00 & \underline{36.36} & \textbf{78.95} & \underline{41.67} & \underline{2.509} \\
InternVL2.5-8B & 0.00 & 0.00 & 0.00 & 30.00 & 4.08 & 8.00 & 18.18 & 57.89 & 13.10 & 1.717 \\
\rowcolor{gray!15}
\textbf{PhyAR (InternVL2.5-8B)} & \textbf{47.06} & \textbf{64.71} & \textbf{33.33} & 50.00 & \textbf{42.86} & 16.00 & \textbf{45.45} & \textbf{78.95} & \textbf{45.83} & \textbf{2.586} \\
\bottomrule
\end{tabular}
\vspace{-14pt}
\end{table*}

For video $\mathcal V$ and query $Q$, we supervise the full sequence $Z=[O;A;Y]$ of observation $O$, attribution $A$, and task-specific verdict $Y$, training the model to carry observed facts through interpretation to judgment. We use the standard autoregressive loss:
\begin{equation}
\mathcal L_{\mathrm{VARC}}
=-\frac{1}{|Z|}\sum_{t=1}^{|Z|}
\log p_\theta(z_t\mid\mathcal V,Q,z_{<t}).
\end{equation}

\subsection{Paired Data Binding}
\label{sec:optimization}

To reduce shortcut learning from shallow scene features, Paired Data Binding treats each matched video pair as a training unit. We use square-root category-balanced sampling and shuffle pairs, keeping each positive--negative pair bound within the same gradient-accumulation window for joint optimization with the VARC loss. Paired videos depict different physical relations or event outcomes in similar scenes, encouraging the model to learn the visual differences that determine event interpretations rather than rely solely on shared scene semantics.

We find that matched pairs exhibit higher gradient cosine similarity $\cos(g^+,g^-)$ than same-category random pairs (Sec.~\ref{sec:ablation}), indicating more aligned local gradient directions and more compatible supervision for joint optimization. Here, $g^+$ and $g^-$ denote gradients of the positive and negative samples' supervised losses with respect to the trainable parameters. PhyAR implements both components through standard supervised fine-tuning with LoRA~\citep{hu2021lora}, adapting different Video-LLM backbones without category-specific architectures.

\vspace{-3pt}
\section{Experiments}
\vspace{-3pt}
\subsection{Experimental Setup}
\vspace{-3pt}

We evaluate paired physical reasoning on PriorPair's 168 test pairs, transfer to the external physical benchmark IntPhys2~\citep{bordes2025intphys2}, and general video understanding on Video-MME~\citep{fu2025videomme}. PriorPair baselines include two proprietary models, GPT-4o~\citep{openai2024gpt4o} and Gemini 2.5 Flash~\citep{comanici2025gemini25}, and eight open-source models: Flash-VStream-7B~\citep{zhang2024flashvstream}, Video-ChatGPT-7B~\citep{maaz2024video}, Video-LLaVA-7B~\citep{lin2024video}, MiniCPM-V 2.6~\citep{yao2024minicpm}, VideoLLaMA2-7B~\citep{cheng2024videollama2}, Qwen3-VL-8B-Instruct~\citep{bai2025qwen3vl}, VideoLLaMA3-7B~\citep{zhang2025videollama3}, and InternVL2.5-8B~\citep{chen2024internvl2}. We apply PhyAR to InternVL2.5-8B (our primary backbone) and VideoLLaMA3-7B to assess architectural compatibility, comparing each trained model with its original backbone to measure training effects. All baselines use released weights or official services under a common test protocol.

We train both PhyAR models on 660 PriorPair training pairs for two epochs with LoRA~\citep{hu2021lora} and square-root category-balanced sampling. PriorPair evaluation uses 16 frames per video, except Video-LLaVA-7B, whose native temporal encoder requires eight. All models use common task-specific prompts and verdict scoring rules, with greedy decoding capped at 512 output tokens. Full training and benchmark-specific evaluation settings appear in Appendix~\ref{appendix:experiments}.

Using the metrics in Sec.~\ref{sec:evaluation_protocol}, we report PCA (\%) as the primary metric. RPB measures the normalized positive--negative accuracy gap. Gemini 3.1 Pro~\citep{google2026gemini31pro} serves as the LLM judge for RAS, rating reasoning quality on a 1--5 scale (full rubric and scoring protocol in Appendix~\ref{appendix:ras_protocol}).

\vspace{-3pt}
\subsection{Main Results on PriorPair}
\vspace{-3pt}

PriorPair poses a substantial challenge to current Video-LLMs (Table~\ref{tab:main_results}). The strongest baseline, Gemini 2.5 Flash, reaches only 32.74\% overall PCA, while the eight open-source Base models score 1.79\%--22.02\%. Some models show localized strengths in Identity Violation and Consequence Arrest but struggle to judge diverse physical phenomena reliably. GPT-4o and Gemini 2.5 Flash outperform all open-source Base models in RAS (2.440 and 2.330, respectively), yet their PCA remains below one-third: stronger physical explanations alone do not ensure correct judgments.

PhyAR leads in overall PCA and RAS. PhyAR--InternVL2.5 reaches 45.83\% PCA and 2.586 RAS, followed by PhyAR--VideoLLaMA3 at 41.67\% and 2.509. PCA rises by 32.73 and 27.98 percentage points over their respective Base models, alongside substantial RAS gains. Without backbone-specific modules, PhyAR improves judgment accuracy and the quality of visually grounded physical reasoning across both streams and all eight categories on two Video-LLMs, demonstrating architectural compatibility. Manual auditing confirmed that sampled RAS ratings followed the rubric, and repeated scoring showed high stability. Validation protocols and results, along with eight paired Base-to-PhyAR improvement examples covering all categories, appear in Appendices~\ref{appendix:ras_audit} and~\ref{appendix:qualitative}, respectively.

\vspace{-3pt}
\subsection{Diagnosing Semantic Prior Dominance}
\label{sec:spd_diagnosis}
\vspace{-3pt}

We analyze Point-wise Accuracy and RPB to diagnose prediction bias. Base models strongly favor prior-aligned predictions, with RPBs of 73.55\% for VideoLLaMA3 and 76.60\% for InternVL2.5 (Table~\ref{tab:spd_diagnosis}a). Positive accuracy exceeds 97\%, but negative accuracy stays below 15\%: judgments reliable under familiar expectations fail when visual evidence contradicts them. PhyAR lowers RPB to 24.56\% and 29.17\%, raising negative accuracy above 50\% on both backbones while maintaining high positive accuracy and improving overall accuracy. PhyAR thus judges prior-conflicting visual content more faithfully while retaining strong performance on prior-aligned counterparts.

We examine how reported visual facts relate to final judgments in InternVL2.5 Base and PhyAR (Table~\ref{tab:spd_diagnosis}b; judging protocol in Appendix~\ref{appendix:spd_protocol}). In Anti-physics, Base observation and verdict accuracy are only 7.09\% and 5.51\%, revealing frequent misreporting of facts indicating physical violations. Despite correct observations, 66.67\% of verdicts rationalize invalid events as plausible. PhyAR improves both accuracies and lowers this conditional error rate to 12.82\%, strengthening factual reporting and judgments grounded in those facts. In Counter-intuitive, both Base accuracies are 36.59\%, yet only 6.67\% of correct observations are followed by wrong verdicts. Once the target event is faithfully described, judgments usually follow visual facts, indicating that failures primarily concern reporting observed events that defy expectations. PhyAR raises both accuracies above 65\%, with no wrong verdicts following correct observations. SPD thus manifests differently across streams: Anti-physics combines omissions of key facts with rationalization despite correctly stated evidence, whereas Counter-intuitive centers on event descriptions shaped by prior expectations.

\begin{table}[t]
  \centering
  \small
  \caption{\textbf{Diagnosing Semantic Prior Dominance.} All values are percentages. (a) Base-to-PhyAR accuracy and RPB. (b) Observation--Verdict diagnosis of InternVL2.5-8B Base and PhyAR on all test negatives. $P(V_{\mathrm{wrong}}\mid O_{\mathrm{correct}})$ is the verdict error rate given correct observations.}
  \label{tab:spd_diagnosis}
  \vspace{-4pt}
  \fontsize{8}{9.5}\selectfont
  \setlength{\tabcolsep}{2pt}
  \begin{minipage}[t]{0.49\linewidth}
    \centering
    \begin{tabular*}{\linewidth}[t]{@{\extracolsep{\fill}}p{\dimexpr0.32\linewidth-4pt\relax}*{4}{>{\centering\arraybackslash}p{\dimexpr0.17\linewidth-4pt\relax}}@{}}
      \toprule
      \multicolumn{5}{c}{\textbf{(a) Base-to-PhyAR Accuracy and RPB}} \\
      \midrule
      \multirow{2}{*}{\textbf{Metric}} & \multicolumn{2}{c}{\textbf{VideoLLaMA3-7B}} & \multicolumn{2}{c}{\textbf{InternVL2.5-8B}} \\
      \cmidrule(lr){2-3}\cmidrule(l){4-5}
      & \textbf{Base} & \textbf{PhyAR} & \textbf{Base} & \textbf{PhyAR} \\
      \midrule
      Positive Acc. $\uparrow$ & 97.62 & 84.52 & 98.81 & 92.26 \\
      Negative Acc. $\uparrow$ & 14.88 & 51.19 & 13.10 & 50.60 \\
      Overall Acc. $\uparrow$ & 56.25 & \textbf{67.86} & 55.95 & \textbf{71.43} \\
      RPB $\downarrow$ & 73.55 & \textbf{24.56} & 76.60 & \textbf{29.17} \\
      \bottomrule
    \end{tabular*}
  \end{minipage}\hfill
  \begin{minipage}[t]{0.49\linewidth}
    \centering
    \setlength{\tabcolsep}{1pt}
    \begin{tabular*}{\linewidth}[t]{@{\extracolsep{\fill}}lcccc@{}}
      \toprule
      \multicolumn{5}{c}{\textbf{(b) Observation--Verdict Diagnosis}} \\
      \midrule
      \multirow{2}{*}{\textbf{Metric}} & \multicolumn{2}{c}{\textbf{Anti-physics}} & \multicolumn{2}{c}{\textbf{Counter-intuitive}} \\
      \cmidrule(lr){2-3}\cmidrule(l){4-5}
      & \textbf{Base} & \textbf{PhyAR} & \textbf{Base} & \textbf{PhyAR} \\
      \midrule
      Observation Acc. $\uparrow$ & 7.09 & \textbf{30.71} & 36.59 & \textbf{65.85} \\
      Verdict Acc. $\uparrow$ & 5.51 & \textbf{44.09} & 36.59 & \textbf{70.73} \\
      $P(V_{\mathrm{wrong}}\mid O_{\mathrm{correct}})\downarrow$ & 66.67 & \textbf{12.82} & 6.67 & \textbf{0.00} \\
      \bottomrule
    \end{tabular*}
  \end{minipage}
\vspace{-10pt}
\end{table}

\vspace{-3pt}
\subsection{Ablation Studies}
\label{sec:ablation}
\vspace{-3pt}

We use supervised fine-tuning without VARC or Paired Data Binding as Baseline SFT. With data and settings fixed, adding either component improves PCA and RAS, and their combination performs best (Table~\ref{tab:ablation_results}a). To isolate pair matching, we retain VARC and randomly bind each positive video to a same-category negative from a different pair. Random-pair training outperforms unpaired training but trails matched-pair training in PCA and RAS: joint positive--negative training is beneficial, with factual contrasts between similar scenes providing additional gains. Our gradient analysis finds higher mean gradient cosine similarity for matched pairs than same-category random pairs in every physical category (overall means: 0.122 versus 0.010; Appendix~\ref{appendix:gradient_cosine}), indicating that matched pairs provide more compatible supervision and better-aligned learning directions during joint optimization.

Using the full PhyAR configuration, we further examine the roles of the two streams in training (Table~\ref{tab:ablation_results}b). Both single-stream configurations improve their own stream but lower the other stream's PCA below Base. Joint training on both streams raises Anti-physics PCA from 25.98\% with Anti-physics-only training to 40.94\%, keeps Counter-intuitive PCA close to its single-stream result, and achieves the highest overall PCA (45.83\%). Thus, under joint training, Counter-intuitive data improve both verification of expected events and judgments of physical violations, showing that the two streams share gains from learning to understand videos through visual evidence.

\begin{table}[t]
  \centering
  \small
  \caption{\textbf{Component and training-stream ablations on PriorPair with InternVL2.5-8B.}}
  \label{tab:ablation_results}
  \vspace{-4pt}
  \fontsize{8}{9.5}\selectfont
  \setlength{\tabcolsep}{1pt}
  \begin{minipage}[t]{0.49\linewidth}
    \centering
    \begin{tabular*}{\linewidth}[t]{@{\extracolsep{\fill}}lcccc@{}}
      \toprule
      \multicolumn{5}{c}{\textbf{(a) Component Ablation}} \\
      \midrule
      \textbf{Configuration} & \textbf{VARC} & \begin{tabular}[c]{@{}c@{}}\textbf{Paired Data}\\\textbf{Binding}\end{tabular} & \begin{tabular}[c]{@{}c@{}}\textbf{Overall}\\\textbf{PCA} $\uparrow$\end{tabular} & \textbf{RAS} $\uparrow$ \\
      \midrule
      Baseline SFT & No & No & 29.17 & 2.137 \\
      w/o VARC & No & Yes & 34.52 & 2.292 \\
      w/o pair binding & Yes & No & 36.31 & 2.330 \\
      Random binding & Yes & Random & 42.86 & 2.470 \\
      Full PhyAR & Yes & Yes & \textbf{45.83} & \textbf{2.586} \\
      \bottomrule
    \end{tabular*}
  \end{minipage}\hfill
  \begin{minipage}[t]{0.49\linewidth}
    \centering
    \begin{tabular*}{\linewidth}[t]{@{\extracolsep{\fill}}lccc@{}}
      \toprule
      \multicolumn{4}{c}{\textbf{(b) Training-stream Ablation (PCA)}} \\
      \midrule
      \multirow{2}{*}{\textbf{Stream}} & \multicolumn{3}{c}{\textbf{Training data}} \\
      \cmidrule(l){2-4}
      & \textbf{Anti-only} & \textbf{Counter-only} & \textbf{Joint} \\
      \midrule
      Anti-physics & 25.98 & 1.57 & \textbf{40.94} \\
      Counter-intuitive & 29.27 & \textbf{63.41} & 60.98 \\
      Overall & 26.79 & 16.67 & \textbf{45.83} \\
      \bottomrule
    \end{tabular*}
  \end{minipage}
\vspace{-5pt}
\end{table}

\subsection{External Generalization and Capability Retention}
\label{sec:external}

We evaluate zero-shot transfer on the external physical benchmark IntPhys2~\citep{bordes2025intphys2}, which tests whether models can distinguish physically possible and impossible events in visually diverse synthetic scenes. We compare Base models using Direct or VARC prompting with PhyAR using VARC prompting (Table~\ref{tab:external_results}a). Direct requests a verdict; VARC requests an Observation--Attribution--Verdict response. VARC prompting alone yields no consistent gains, whereas PhyAR achieves the highest overall accuracy on both backbones: 51.09\% for InternVL2.5 and 52.27\% for VideoLLaMA3. Meanwhile, PhyAR substantially improves judgment accuracy on impossible events and markedly reduces the models' bias toward possible events. These results demonstrate that PhyAR's reduction of prior-driven judgment bias generalizes to external physical tasks.

We assess general video understanding on Video-MME~\citep{fu2025videomme} using its native visual-only multiple-choice protocol, without VARC prompting for Base or PhyAR (Table~\ref{tab:external_results}b). Overall accuracy changes from 61.41\% to 61.93\% on InternVL2.5 and from 60.30\% to 60.15\% on VideoLLaMA3. PhyAR largely preserves general video understanding. See Appendices~\ref{appendix:intphys2_evaluation} and~\ref{appendix:videomme_evaluation}.

\begin{table*}[t]
  \centering
  \small
  \caption{\textbf{External generalization and capability retention.} (a) IntPhys2 zero-shot accuracy (\%); Class Gap: absolute class accuracy difference (percentage points). (b) Video-MME accuracy (\%).}
  \label{tab:external_results}
  \vspace{-4pt}
  \fontsize{8}{9.5}\selectfont
  \setlength{\tabcolsep}{0.5pt}
  \begin{minipage}[t]{0.62\linewidth}
    \centering
    \begin{tabular*}{\linewidth}[t]{@{\extracolsep{\fill}}lccccc@{}}
      \toprule
      \multicolumn{6}{c}{\textbf{(a) IntPhys2}} \\
      \midrule
      \textbf{Model} & \textbf{Prompt} & \begin{tabular}[c]{@{}c@{}}\textbf{Overall}\\\textbf{Acc.} $\uparrow$\end{tabular} & \begin{tabular}[c]{@{}c@{}}\textbf{Possible}\\\textbf{Acc.} $\uparrow$\end{tabular} & \begin{tabular}[c]{@{}c@{}}\textbf{Impossible}\\\textbf{Acc.} $\uparrow$\end{tabular} & \begin{tabular}[c]{@{}c@{}}\textbf{Class}\\\textbf{Gap} $\downarrow$\end{tabular} \\
      \midrule
      InternVL2.5-8B & Direct & 50.20 & 93.48 & 6.92 & 86.56 \\
      InternVL2.5-8B & VARC & 49.11 & 85.57 & 12.65 & 72.92 \\
      PhyAR (InternVL2.5-8B) & VARC & \textbf{51.09} & 45.45 & 56.72 & \textbf{11.26} \\
      \midrule
      VideoLLaMA3-7B & Direct & 49.01 & 80.24 & 17.79 & 62.45 \\
      VideoLLaMA3-7B & VARC & 49.80 & 83.00 & 16.60 & 66.40 \\
      PhyAR (VideoLLaMA3-7B) & VARC & \textbf{52.27} & 54.74 & 49.80 & \textbf{4.94} \\
      \bottomrule
    \end{tabular*}
  \end{minipage}\hfill
  \begin{minipage}[t]{0.36\linewidth}
    \centering
    \renewcommand{\arraystretch}{1.3}
    \begin{tabular*}{\linewidth}[t]{@{\extracolsep{\fill}}lc@{}}
      \toprule
      \multicolumn{2}{c}{\textbf{(b) Video-MME}} \\
      \midrule
      \textbf{Model} & \begin{tabular}[c]{@{}c@{}}\textbf{Overall}\\\textbf{Acc.} $\uparrow$\end{tabular} \\
      \midrule
      InternVL2.5-8B & 61.41 \\
      PhyAR (InternVL2.5-8B) & \textbf{61.93} \\
      \midrule
      VideoLLaMA3-7B & \textbf{60.30} \\
      PhyAR (VideoLLaMA3-7B) & 60.15 \\
      \bottomrule
    \end{tabular*}
  \end{minipage}
\vspace{-10pt}
\end{table*}

\paragraph*{Conclusion and Limitations.}

We identify Semantic Prior Dominance in Video-LLMs' physical reasoning: models rationalize physically impossible events and incorrectly report expected outcomes despite conflicting visual evidence in physically plausible, counter-intuitive scenarios. PriorPair unifies these failures in a paired benchmark, while PhyAR combines VARC and Paired Data Binding to ground physical reasoning in visual evidence. Across two backbones, this lightweight framework improves judgment accuracy and reasoning quality, reduces prior-driven bias, and transfers to an external physical benchmark while largely preserving general video understanding.

PriorPair prioritizes high fidelity and physical rigor, but expert verification costs limit its scale. Future work will explore automated construction and annotation to expand the dataset while retaining expert quality control.

\bibliographystyle{iclr2027_conference}
\bibliography{refs}

\clearpage
\appendix
\raggedbottom
\begin{center}
{\Large\bfseries Appendix for}\\[4pt]
{\large\itshape From Priors to Perception: Grounding Video-LLMs in Physical Reality}
\end{center}
\vspace{0.5em}

\section{PriorPair Dataset Details}
\label{appendix:taxonomy}

PriorPair is organized into two streams, four groups, and eight categories according to temporal continuity, object permanence, physical interaction constraints, and target-event occurrence. Anti-physics positive and negative videos depict physically valid and invalid processes, respectively, and the task is physical-validity verification. Both videos in a Counter-intuitive pair remain physically valid, but the same target event occurs only in the positive video; the task is target-event occurrence verification. Paired videos share similar scene semantics but differ in the temporal relations, object states, motion, or interaction outcomes that determine the verdict. Each video is accompanied by complete Observation--Attribution--Verdict (O--A--V) annotations.

\subsection{Stream 1: Anti-physics}
\label{appendix:taxonomy_antiphysics}

\subsubsection*{Temporal Fallacies}

\paragraph{(1) Coherence Violation.}
\begin{itemize}
    \item \textbf{Definition:} This category disrupts the local temporal continuity of actions. Positions, poses, or action stages in adjacent clips lack continuous motion connecting them, causing objects to abruptly jump, reset, or switch discontinuously between stages within the same scene.

    \item \textbf{Examples:}
For example, a car originally travels continuously along a straight road, but shuffled clips make it abruptly jump back and forth between road positions. Reordering walking clips can also produce a \lq\lq{}left foot--right foot--left foot--left foot--right foot\rq\rq{} sequence without the necessary transitions, accompanied by discontinuities in body position and leg pose. In a video of a person descending wooden steps toward the camera, the positive video preserves the continuous descent, whereas the negative video abruptly switches between the top of the steps, a close-up near the camera, and the middle of the descent.
\end{itemize}

\paragraph{(2) Causal Reversal.}
\begin{itemize}
    \item \textbf{Definition:} This category inverts the order of causes and effects with explicit physical dependencies, presenting an outcome before the corresponding action. Relevant event stages include force application and fracture, collision and displacement, and pouring and liquid accumulation.

    \item \textbf{Examples:}
For example, a cup is thrown, strikes the floor, and then shatters; the negative video instead presents shattering on the floor before a person throws the intact cup. Similarly, a vase shattering before it begins to fall reverses the order of falling, impact, and fracture. A positive video of clamps pulling a green sheet presents force application, deformation, and tearing in sequence, whereas the negative video shows tearing first and then cuts back to the clamps pulling the intact sheet. Liquid accumulating at the bottom of a receiving container before pouring begins is another example of causal reversal.
\end{itemize}

\subsubsection*{Permanence Fallacies}

\paragraph{(3) Existence Violation.}
\begin{itemize}
    \item \textbf{Definition:} This category makes objects or matter appear or disappear without plausible creation, removal, transport, or transformation, including changes in unoccluded regions and changes occurring during occlusion.

    \item \textbf{Examples:}
\begin{itemize}
    \item Typical cases include an object suddenly appearing on an empty table; a stationary object disappearing without leaving, being removed, or undergoing a physical transformation; a previously covered object being absent after uncovering; and a new object appearing beneath a cover that originally concealed an empty region. For example, a basket descends to cover a ball on a table. The positive video reveals the same ball when the basket is lifted, whereas the negative video reveals an empty tabletop.

    \item Other examples include a soft cloth suddenly collapsing flat after covering an object, with no path for the object to leave; removing a small cloth to reveal a large object that could not have been fully concealed beneath it, with no alternative entry path; and a receiving container's liquid level continuously rising even though no liquid flows from the pouring position and no other supply is present.
\end{itemize}
\end{itemize}

\paragraph{(4) Identity Violation.}
\begin{itemize}
    \item \textbf{Definition:} This category abruptly changes a continuously tracked object's identity, shape, quantity, or intrinsic properties without a plausible physical transition.

    \item \textbf{Examples:}
\begin{itemize}
    \item For example, a quail suddenly becomes a rooster while walking, or a dog becomes a sheep or cat while following its original path and maintaining its position and movement rhythm.

    \item Quantity changes include one ball becoming two or more complete balls after landing, without a plausible replication or splitting process or additional objects entering the scene.

\end{itemize}
\end{itemize}

\subsubsection*{Interaction Fallacies}

\paragraph{(5) Dynamic Violation.}
\begin{itemize}
    \item \textbf{Definition:} This category makes motion, force balance, or thermodynamic evolution inconsistent with the physical conditions of the scene. It encompasses inertia and momentum anomalies, trajectory anomalies, missing driving mechanisms, equilibrium anomalies, and anomalous processes produced by continuous reverse playback.

    \item \textbf{Examples:}
\begin{itemize}
    \item \textit{Inertia, friction, and momentum anomalies.}
    An object sliding on a smooth surface instantly stops or rebounds despite the absence of an obstacle, braking, or another significant interaction. An object moving on a rough carpet after the push ends accelerates rather than slows down, or maintains constant speed over a distance where friction is significant. Collision examples include a moving ball striking a stationary ball and both immediately stopping, with no other object or external interaction absorbing the corresponding momentum. Energy anomalies include a basketball rebounding above its initial release height after a passive collision, without additional energy input.

    \item \textit{Trajectory and gravity anomalies.}
    A basketball or stone continues along a straight diagonal path without downward deflection when gravity dominates and no additional support or propulsion is present. Other examples include an object following an S-shaped falling path under gravity-dominated conditions and a poured stream maintaining an S-shaped trajectory in mid-air without lateral driving. Falling behavior can also conflict with material and air-resistance conditions: a dense stone slowly drifts downward without sufficient upward force, or a feather falls at a speed incompatible with its mass and projected area under conditions of significant air resistance. Further examples include a falling object suddenly hovering and an object pushed off a table stopping in mid-air without a supporting structure.

    \item \textit{Missing driving mechanisms.}
    A stationary toy duck starts moving and accelerates across a level tabletop without a push or another driving mechanism. After the acting agent is removed, a ball rolls, liquid stirs, a door opens, a pen writes, or an object rises on its own. These scenes contain no actuator, flow, tilt, or other mechanism sufficient to explain the motion.

    \item \textit{Anti-gravity and anomalous reverse evolution.}
    Fragments spontaneously assemble into an intact object, spilled liquid gathers and returns to its original container, or fallen dominoes successively stand upright without external manipulation or the required energy input. Water continuously flowing uphill without a pressure gradient, pumping, capillary action, or another mechanism is another example.

    \item \textit{Force-balance anomalies.}
    The more heavily loaded end of an equal-arm seesaw remains above the other end without an additional force, support, or fixing structure. A block tower remains suspended after losing the support necessary for stability. Loose objects remain on a plate tilted beyond the angle at which they can stay in place, without adhesion, fixing, or an obstructing structure.
\end{itemize}
\end{itemize}

\paragraph{(6) Constraint Violation.}
\begin{itemize}
    \item \textbf{Definition:} This category breaches spatial boundaries, material properties, conservation relations, or physical constraints linking interaction conditions to outcomes. It encompasses environmental-response mismatches, disproportionate interaction responses, anomalous material behavior, and failures of rigid-body or medium interaction constraints.

    \item \textbf{Examples:}
\begin{itemize}
    \item \textit{Environmental and thermodynamic response mismatches.}
    An ordinary candle directly exposed to strong airflow maintains the same steady flame as in still air. Ice or butter receives sufficiently intense and prolonged heating but undergoes no expected phase change. Ordinary fuel continues burning without oxygen or another oxidizer. A wood-fueled campfire fully submerged in water continues to display bright flames without an additional oxygen supply or a special fuel mechanism. Environmental-motion examples include flags or hair remaining completely still while directly exposed to airflow that visibly bends surrounding trees, without shielding, fixing, or another explanation.

    \item \textit{Disproportionate interaction responses.}
    A gently falling feather or a small, low-speed stone produces a splash far exceeding its available energy. A small stone enters water under conditions sufficient to create visible disturbance, but the surface shows no response. Other examples include a small stone producing a water jet approximately fifty meters high and a massive boulder entering water at a speed sufficient to disturb it while the surface remains completely flat.

    \item \textit{Conservation and mixing anomalies.}
    A cup's liquid quantity remains unchanged while an appreciable amount is continuously consumed, without replenishment. A cup's liquid level rises even though no liquid leaves the kettle. Sand continuously flows downward through an hourglass, but the amount in its upper chamber does not decrease. Known miscible substances remain completely unmixed after sufficient contact and stirring, or the resulting color, form, or material state cannot be explained by the inputs and their reaction processes.

    \item \textit{Anomalous material responses.}
    An ordinary rigid vase exhibits rubber-like compression, elastic rebound, and shape recovery upon impact, without material replacement or a process that changes its properties. A paper hammer easily shatters intact, sturdy glass under impact conditions insufficient to break it. Substances being stirred respond in ways incompatible with their known hardness, fluidity, or state. When a knife cuts an ordinary orange, its soft, juicy flesh abruptly undergoes anomalous brittle fragmentation into more than six pieces without material conditions supporting that response.

    \item \textit{Spatial-boundary and medium-interaction anomalies.}
    A robotic arm passes directly through an intact solid wall, or two solid spheres pass through one another after collision without corresponding displacement, damage, or deformation. A spoon, knife, or hand enters a solid or liquid and occupies space that should produce a visible response, but the target undergoes no corresponding deformation, displacement, or movement.
\end{itemize}
\end{itemize}

\subsection{Stream 2: Counter-intuitive}
\label{appendix:taxonomy_counterintuitive}

\subsubsection*{Counter-intuitive Events}

\paragraph{(7) Near-Miss.}
\begin{itemize}
    \item \textbf{Definition:} In this category, an object approaches a potential interaction region, but a visible spatial gap or a halt before contact prevents the target interaction. The positive video establishes the interaction, whereas the negative video does not; both remain physically valid.

Near-Miss also includes object transfer or action completion that depends on spatial alignment and the relevant contact. Non-completion in these cases arises from a gap between the relevant positions, non-intersecting trajectories, or a halt before contact, rather than an arbitrary unfinished action or a missing downstream consequence after the target interaction has been established.

    \item \textbf{Examples:}
\begin{itemize}
    \item \textit{Trajectory Miss.}
    An object passes beside the target, and its trajectory does not intersect the relevant contact region or opening. For example, a ball rolls toward dominoes but passes beside them, leaving them upright; a foot swings toward a soccer ball but passes beside it, leaving the ball stationary; or a golf ball struck by a club rolls alongside a hole without entering it. The corresponding positive videos show the ball contacting the dominoes, the foot contacting the soccer ball, and the golf ball entering the hole, respectively.

    \item \textit{Abrupt Halt.}
    An approaching motion ends before the relevant contact, leaving a visible gap. For example, a finger stops before touching the first domino, which remains unmoved, or a hammer stops before striking glass, which remains intact. In the corresponding positive videos, the approach continues until the gap closes and the target contact is established.
\end{itemize}
\end{itemize}

\paragraph{(8) Consequence Arrest.}
\begin{itemize}
    \item \textbf{Definition:} In this category, an initiating action or contact has occurred, but its magnitude, angle, displacement, pressure, duration, or other conditions are insufficient to produce the specified consequence. The positive video shows the target consequence, whereas the negative video does not; both remain physically valid. A preparatory pose alone does not establish that the initiating action has occurred.

    \item \textbf{Examples:}
\begin{itemize}
    \item \textit{Insufficient Impact.}
    Contact occurs but is insufficient to cause the target displacement, fracture, collapse, or chain reaction. For example, a hammer contacts glass without enough impact to break it; an agent contacts and pushes a box, but the box does not move; or a tool touches the first domino without destabilizing it, leaving the remaining dominoes upright. In a positive video of a cup falling onto a concrete block, the cup shatters; in the negative video, it remains intact and rebounds from the contact surface.

    \item \textit{Threshold Failure.}
    An action occurs but does not reach the conditions required to trigger the target state change. For example, a hand tilts a water cup, but insufficient tilt, a liquid surface remaining below the rim, or surface tension prevents spilling. The corresponding positive video shows the tilt carrying liquid over the rim. Other cases involve insufficient action amplitude, pressure, or duration to trigger a specified deformation or subsequent motion.
\end{itemize}
\end{itemize}

In Near-Miss, a spatial gap or a halt before contact prevents the target interaction from being established. In Consequence Arrest, the initiating action or contact has occurred, but the specified consequence is absent. The latter differs from Constraint Violation because insufficient interaction magnitude or other physical conditions can explain the missing consequence, rather than creating a physical contradiction between conditions and responses.

\begin{figure}[t]
    \centering
    \includegraphics[width=\linewidth]{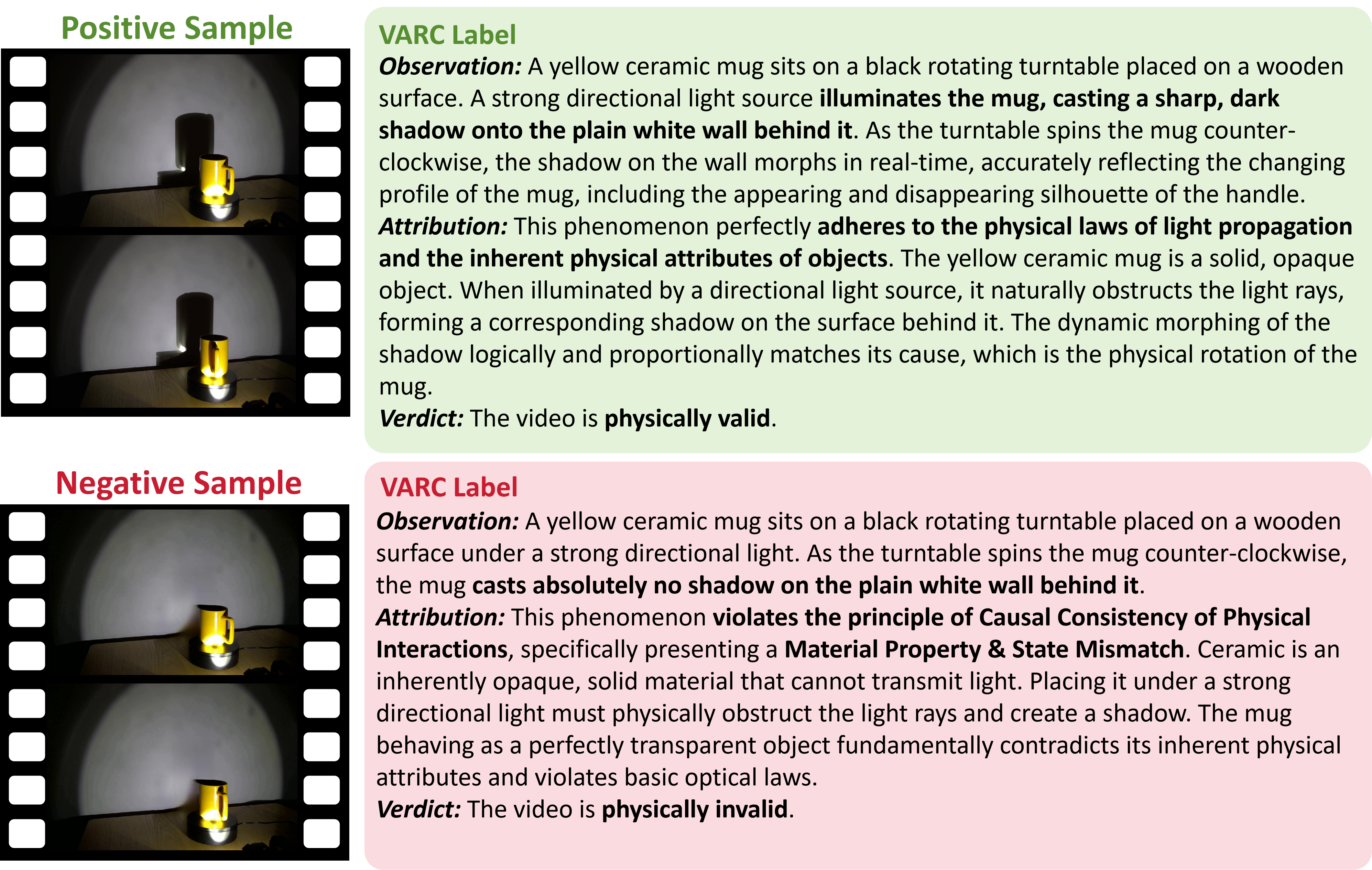}
    \caption{\textbf{A Constraint Violation pair and its O--A--V annotations.} Under directional illumination, the positive video shows the opaque mug casting a shadow on the wall, whereas the negative video lacks this shadow under corresponding scene conditions. The annotations describe the visual difference, attribute it to light propagation and material constraints, and give physically valid and physically invalid verdicts, respectively.}
    \label{fig:varc_constraint_example}
\end{figure}

Figure~\ref{fig:varc_constraint_example} presents a Constraint Violation pair with complete O--A--V annotations. In the positive video, the mug casts a shadow that changes with its rotation; in the negative video, the mug remains in the light's path but its shadow on the wall is absent.

\subsection{Dataset Size and Category Distribution}

PriorPair contains 828 matched video pairs, with 660 training pairs and 168 test pairs. These comprise 1,656 video sample records referencing 1,616 unique video files. Videos have an average duration of 4.0 seconds. Table~\ref{tab:priorpair_category_distribution} summarizes the category counts and splits.

\begin{table}[t]
    \centering
    \caption{Category distribution of PriorPair. Counts refer to matched video pairs.}
    \label{tab:priorpair_category_distribution}
    \small
    \setlength{\tabcolsep}{4pt}
    \begin{tabular}{llrrrr}
        \toprule
        Group & Category & Train & Test & Total & Share \\
        \midrule
        \multirow{2}{*}{Temporal Fallacies}
        & Coherence Violation & 67 & 17 & 84 & 10.14\% \\
        & Causal Reversal & 67 & 17 & 84 & 10.14\% \\
        \midrule
        \multirow{2}{*}{Permanence Fallacies}
        & Existence Violation & 34 & 9 & 43 & 5.19\% \\
        & Identity Violation & 38 & 10 & 48 & 5.80\% \\
        \midrule
        \multirow{2}{*}{Interaction Fallacies}
        & Dynamic Violation & 196 & 49 & 245 & 29.59\% \\
        & Constraint Violation & 97 & 25 & 122 & 14.73\% \\
        \midrule
        \multirow{2}{*}{Counter-intuitive Events}
        & Near-Miss & 88 & 22 & 110 & 13.29\% \\
        & Consequence Arrest & 73 & 19 & 92 & 11.11\% \\
        \midrule
        Total & & 660 & 168 & 828 & 100\% \\
        \bottomrule
    \end{tabular}
\end{table}

The Anti-physics stream contains 626 pairs, and the Counter-intuitive stream contains 202 pairs. Each pair presents corresponding versions of the core physical process or target event in semantically matched scenes, supporting joint evaluation of positive and negative judgments.

\subsection{Construction and Quality Control}
\label{appendix:construction_quality}

PriorPair follows a four-stage human-in-the-loop construction pipeline. Source videos come from Something-Something V2, PhysBench, Panda-70M, ImplausiBench, ActivityNet, Physics-IQ, and Pexels. Gemini 3.1 Pro is used for visual fact extraction, target-scenario and generation-prompt design, and O--A--V annotation generation. Video generation models include Kling and Veo.

\subsubsection*{Stage 1: Selection \& Clipping}
Experts select videos with clear subjects and key processes and trim them into short clips centered on a single physical interaction. The predefined taxonomy guides the choice of positive anchors, target categories, and counterpart scenarios. Counter-intuitive pairs share a short, atomic, directly observable target event, such as \lq\lq{}The foot contacts the soccer ball\rq\rq{} or \lq\lq{}The cup shatters.\rq\rq{}

\subsubsection*{Stage 2: Visual Fact Anchoring}
Gemini 3.1 Pro describes visible subjects, actions, environments, spatial relations, contacts, and state changes, and extracts visual style and camera motion information. Experts check the descriptions, remove objects, actions, and outcomes absent from the video, and exclude physical inferences from the observations. Visual fact anchors record observable changes such as downward motion, without incorporating causal explanations such as \lq\lq{}due to gravity\rq\rq{} into the Observation.

\subsubsection*{Stage 3: Counterpart Construction and Annotation}
Given the positive anchor, verified visual facts, and target scenario, the pipeline selects deterministic manipulation or generative synthesis.

Deterministic manipulation reuses the original video content for scenarios achievable through temporal rearrangement or localized modification. Coherence Violation divides a video into three to five clips and shuffles their order. Causal Reversal exchanges cause and effect clips. Some Dynamic Violation videos use continuous reverse playback to produce anomalous processes. Experts perform localized removal and content restoration with assistance from SAM 2 and ProPainter.

Generative synthesis is used for scenarios requiring new motion, altered trajectories, object transformations, or reconstructed interaction responses. Target-specific prompts describe the required visible changes while preserving the subjects, environment, camera, and visual style of the paired videos as closely as possible. Anti-physics prompts specify the intended physical violation. Counter-intuitive prompts preserve physical validity and change only the occurrence status of the shared target event. Complete scenario-construction, video-generation, and annotation prompts are provided in Appendices~\ref{appendix:prompt_scenario}--\ref{appendix:prompt_annotation}.

Gemini 3.1 Pro generates O--A--V annotations for each video. Negative annotations are formed from verified positive descriptions and annotations together with the counterpart scenario, and are then reviewed and corrected by experts against the final negative video. Observation records visible facts, Attribution explains their relation to the relevant physical conditions or target event, and Verdict gives the task-specific conclusion. Anti-physics verdicts are \lq\lq{}The video is physically valid.\rq\rq{} or \lq\lq{}The video is physically invalid.\rq\rq{} Counter-intuitive verdicts are \lq\lq{}The target event occurred.\rq\rq{} or \lq\lq{}The target event did not occur.\rq\rq{}

\subsubsection*{Stage 4: Final Verification}
Three experts independently review all 828 pairs for visual quality, target phenomena, counterpart correspondence, category assignments, O--A--V annotations, and unintended visual cues. Verification requires the target phenomenon to be clearly visible in the final videos, the annotations to match the video facts, and the paired videos to contain no additional process that changes the interpretation of the target phenomenon. For Counter-intuitive pairs, experts further check that both videos remain physically valid, share the same target event, and depict its occurrence and non-occurrence, respectively.

Through category-directed construction, semantically matched pairing, and comprehensive expert review, PriorPair focuses the task on the visual differences that determine event interpretation. Its taxonomy, editing and generation routing, and prompt templates provide a reusable construction foundation for future expansion.

\section{Experimental and Training Details}
\label{appendix:experiments}

\subsection{Training Setup}
\label{appendix:training_setup}

PhyAR is trained from the InternVL2.5-8B and VideoLLaMA3-7B base models. The training set contains 660 original pairs and is square-root balanced to 985 pairs per epoch. Both models are trained for two epochs. The cosine learning-rate schedule is parameterized over a three-epoch horizon with a 5\% linear warmup.

Both backbones use 4-bit NormalFloat (NF4) quantization with double quantization, \texttt{bfloat16} computation, and LoRA fine-tuning. LoRA rank, scaling factor, and dropout are set to 16, 32, and 0.1, respectively, and LoRA is applied to the language model's attention and feed-forward projections and the multimodal projector. The vision encoder and all remaining parameters are frozen. Training uses 8-bit AdamW with an initial learning rate of $5\times10^{-5}$ on four NVIDIA RTX A6000 GPUs. Training takes approximately 3 hours 04 minutes for InternVL2.5-8B and 5 hours 54 minutes for VideoLLaMA3-7B.
The training configurations are summarized in Table~\ref{tab:hyperparams}.

\begin{table*}[t]
\centering
\caption{\textbf{Training configurations for PhyAR.}}
\label{tab:hyperparams}
\resizebox{\textwidth}{!}{%
\begin{tabular}{lcc}
\toprule
\textbf{Setting} & \textbf{InternVL2.5-8B} & \textbf{VideoLLaMA3-7B} \\
\midrule
Original training pairs & \multicolumn{2}{c}{660} \\
Balanced pairs per epoch & \multicolumn{2}{c}{985} \\
Frames per video & \multicolumn{2}{c}{16} \\
Training epochs & \multicolumn{2}{c}{2} \\
Quantization & \multicolumn{2}{c}{4-bit NF4 with double quantization} \\
LoRA configuration & \multicolumn{2}{c}{$r=16$, $\alpha=32$, dropout $=0.1$} \\
Optimizer & \multicolumn{2}{c}{8-bit AdamW} \\
Learning rate & \multicolumn{2}{c}{$5\times10^{-5}$} \\
GPUs & \multicolumn{2}{c}{4 $\times$ NVIDIA RTX A6000} \\
Trainable parameters & 38,010,880 & 40,560,640 \\
Training time & 3 h 04 min & 5 h 54 min \\
\bottomrule
\end{tabular}%
}
\end{table*}

\subsection{Category Balancing and Paired Data Binding}
\label{appendix:category_balancing_pdb}

Let $n_c$ be the number of original training pairs in category $c$ and $n_{\max}=196$ the size of the largest category. The target count for category $c$ in each epoch is
\begin{equation}
\hat n_c=\left\lceil\sqrt{n_c n_{\max}}\right\rceil.
\end{equation}
All original pairs are retained, and the remaining quota is filled by repeated sampling within each category. This produces targets of 115, 115, 82, 87, 196, 138, 132, and 120 pairs for Coherence, Causal Reversal, Existence, Identity, Dynamic, Constraint, Near-Miss, and Consequence Arrest, respectively, totaling 985 pairs per epoch.

Paired Data Binding organizes training around complete positive--negative pairs, ensuring that both videos in a pair contribute to the same parameter update. Pairs are shuffled while their internal correspondence is preserved. This mechanism introduces no additional contrastive loss and retains the autoregressive VARC objective defined in the main paper.

\subsection{Controlled Ablation Settings}
\label{appendix:ablation_settings}

All ablations use InternVL2.5-8B and retain the same frame sampling, category balancing, LoRA configuration, optimizer, learning rate, and training schedule as the full model.

\begin{itemize}
    \item \textbf{Baseline SFT} removes both VARC and Paired Data Binding, globally shuffles individual videos, and uses only task-specific verdict supervision.
    \item \textbf{w/o VARC} retains matched-pair binding but replaces O--A--V supervision with direct verdict supervision.
    \item \textbf{w/o Paired Data Binding} retains VARC but globally shuffles individual videos without preserving their positive--negative correspondence.
    \item \textbf{Random Binding} retains VARC and local positive--negative binding but pairs each positive video with a same-category negative from another original pair. The mapping excludes true counterparts and is regenerated every epoch.
\end{itemize}

The stream ablations retain only the six Anti-physics categories or the two Counter-intuitive categories while preserving the remaining training settings.

\subsection{PriorPair Evaluation}
\label{appendix:priorpair_evaluation}

The PriorPair test set contains 168 pairs and 336 videos across eight categories. We evaluate GPT-4o, Gemini 2.5 Flash, Flash-VStream-7B, Video-ChatGPT-7B, Video-LLaVA-7B, MiniCPM-V 2.6, VideoLLaMA2-7B, Qwen3-VL-8B-Instruct, VideoLLaMA3-7B, InternVL2.5-8B, and the two PhyAR models.

All models use greedy decoding with a maximum of 512 output tokens. Sixteen frames are uniformly sampled from each video, except for Video-LLaVA-7B, which uses eight frames. The Anti-physics task asks whether the video is physically valid, whereas the Counter-intuitive task asks whether the specified target event occurred. For Flash-VStream, Video-ChatGPT, Video-LLaVA, VideoLLaMA2, and VideoLLaMA3 Base, an identical format constraint is appended to require Observation, Attribution, and Verdict in order without changing the task or verdict options. Complete task prompts are provided in Appendix~\ref{appendix:prompt_tasks}.

Predictions are parsed from the final Verdict. Unparseable outputs are counted as incorrect, and their corresponding pairs fail PCA. The output-parsing procedure and the PCA, RPB, and RAS metrics are detailed in Appendix~\ref{appendix:ras_protocol}.

\subsection{IntPhys2 Evaluation}
\label{appendix:intphys2_evaluation}

Zero-shot transfer is evaluated on the complete IntPhys2 Main set without using IntPhys2 data for training. The set contains 506 Possible and 506 Impossible videos across continuity (240), immutability (240), permanence (240), and solidity (292). All models use 16 uniformly sampled frames and greedy decoding.

The \textbf{Direct} setting uses the official IntPhys2 yes/no prompt with at most 32 output tokens, whereas the \textbf{VARC} setting uses the physical-validity O--A--V prompt with at most 512 output tokens. Direct predictions are parsed from the final unambiguous yes/no response, whereas VARC predictions are parsed from the final physical-validity Verdict. Unparseable outputs are counted as incorrect. Overall Accuracy is computed over all 1,012 videos, and Class Gap is the absolute difference between Possible and Impossible accuracy.

\subsection{Video-MME Evaluation}
\label{appendix:videomme_evaluation}

Video-MME is evaluated in the visual-only setting without subtitles. The benchmark contains 2,700 multiple-choice questions over 900 videos, with 900 questions in each of the Short, Medium, and Long groups. Base and PhyAR models use the original multiple-choice prompt without VARC, uniformly sample 16 frames, apply greedy decoding, and generate at most eight output tokens. Predictions are parsed from an isolated option letter and otherwise by matching the generated response to an answer option. Unparseable outputs are counted as incorrect. Detailed results by video duration are shown in Table~\ref{tab:videomme_detailed}.

\begin{table}[t]
\centering
\caption{\textbf{Detailed visual-only Video-MME accuracy (\%).}}
\label{tab:videomme_detailed}
\resizebox{\columnwidth}{!}{%
\begin{tabular}{lcccc}
\toprule
\textbf{Model} & \textbf{Overall} & \textbf{Short} & \textbf{Medium} & \textbf{Long} \\
\midrule
InternVL2.5-8B & 61.41 & 73.33 & 59.33 & 51.56 \\
PhyAR (InternVL2.5-8B) & \textbf{61.93} & \textbf{73.89} & \textbf{60.11} & \textbf{51.78} \\
\midrule
VideoLLaMA3-7B & \textbf{60.30} & \textbf{74.22} & \textbf{56.56} & 50.11 \\
PhyAR (VideoLLaMA3-7B) & 60.15 & \textbf{74.22} & 55.89 & \textbf{50.33} \\
\bottomrule
\end{tabular}%
}
\end{table}

\section{Reasoning Alignment Score Evaluation and Validation}

\subsection{RAS Scoring Protocol}
\label{appendix:ras_protocol}

The \textbf{Reasoning Alignment Score (RAS)} uses Gemini 3.1 Pro to evaluate the overall alignment between a model response and its ground-truth Observation--Attribution--Verdict annotation. The evaluator first determines whether the task-specific Verdict is correct and then assesses the alignment of the Observation and Attribution. An incorrect Verdict restricts the score to 1--2, whereas a correct Verdict restricts it to 3--5.

\begin{table}[t]
\centering
\small
\caption{\textbf{Hierarchical RAS scoring rubric.}}
\label{tab:ras_rubric}
\begin{tabular}{c p{0.79\columnwidth}}
\toprule
\textbf{Score} & \textbf{Criterion} \\
\midrule
1 & The Verdict is incorrect or missing, and the Observation and Attribution are predominantly irrelevant, severely hallucinated, or fundamentally inconsistent. \\
2 & The Verdict is incorrect or missing, but some relevant entities, actions, visual evidence, or constraints are correctly identified. \\
3 & The Verdict is correct, but the Observation or Attribution is missing, predominantly incorrect, or unsupported. A bare correct Verdict also receives 3. \\
4 & The Verdict is correct and the Observation and Attribution are substantially aligned, with only minor errors or non-decisive omissions. \\
5 & The Verdict is correct, the decisive visual facts are accurate, and the Attribution correctly connects the evidence to the conclusion without material errors or omissions. \\
\bottomrule
\end{tabular}
\end{table}

For pair $i$, let $s_i^+$ and $s_i^-$ denote the positive and negative sample scores, and let $a_i^+,a_i^-\in\{0,1\}$ denote their Verdict correctness. Strict pair-wise RAS is
\begin{equation}
r_i=
\begin{cases}
\dfrac{s_i^+ + s_i^-}{2}, & a_i^+=a_i^-,\\[5pt]
s_i^+, & a_i^+=0,\ a_i^-=1,\\
s_i^-, & a_i^+=1,\ a_i^-=0.
\end{cases}
\end{equation}
Thus, when only one Verdict is incorrect, the pair score is determined by the incorrect counterpart; otherwise, the two scores are averaged. Category-level RAS averages $r_i$ within each category, and Overall RAS is the micro-average over all 168 test pairs. Predictions are parsed from the final task-specific Verdict, with unparseable outputs counted as incorrect; Point-wise Accuracy, PCA, and RPB then follow the definitions in Sec.~\ref{sec:evaluation_protocol}. Complete evaluator prompts are provided in Appendix~\ref{appendix:prompt_ras}.

\subsection{Category-wise RAS Results}
\label{appendix:ras_results}

\begin{table*}[t]
\centering
\small
\setlength{\tabcolsep}{4.2pt}
\renewcommand{\arraystretch}{0.95}
\caption{\textbf{Category-wise RAS (1--5) on PriorPair.} Overall RAS is micro-averaged over all 168 test pairs using strict pair-wise aggregation. Category abbreviations and highlighting conventions follow Table~\ref{tab:main_results}.}
\label{tab:ras_category_results}
\vspace{-4pt}
\fontsize{8}{9.5}\selectfont
\begin{tabular}{l cccccc cc c}
\toprule
\multirow{2}{*}{\textbf{Method}} & \multicolumn{6}{c}{\textbf{Anti-physics}} & \multicolumn{2}{c}{\textbf{Counter-intuitive}} & \multirow{2}{*}{\textbf{Overall}$\uparrow$} \\
\cmidrule(lr){2-7} \cmidrule(lr){8-9}
 & \textbf{Coh.} & \textbf{Cau.} & \textbf{Exi.} & \textbf{Ide.} & \textbf{Dyn.} & \textbf{Cns.} & \textbf{Near-Miss} & \textbf{C-Arrest} & \\
\midrule
\multicolumn{10}{c}{\textit{Proprietary Models}} \\
\midrule
GPT-4o & 2.147 & 1.735 & 1.667 & \textbf{4.150} & \underline{2.204} & \textbf{2.480} & 2.432 & 3.368 & 2.440 \\
Gemini 2.5 Flash & 1.824 & 2.382 & 1.444 & \underline{3.450} & \textbf{2.316} & \underline{2.400} & 2.091 & 2.789 & 2.330 \\
\midrule
\multicolumn{10}{c}{\textit{Open-Source Models}} \\
\midrule
Flash-VStream-7B & 1.118 & 1.176 & 1.333 & 1.400 & 1.194 & 1.360 & 1.182 & 1.395 & 1.250 \\
Video-ChatGPT-7B & 1.000 & 1.000 & 1.000 & 1.000 & 1.020 & 1.000 & 1.477 & 1.289 & 1.101 \\
Video-LLaVA-7B & 1.294 & 1.176 & 1.222 & 1.950 & 1.082 & 1.420 & 1.205 & 1.526 & 1.289 \\
MiniCPM-V 2.6 & 1.500 & 1.000 & 1.000 & 1.550 & 1.122 & 1.280 & 1.795 & 3.237 & 1.518 \\
VideoLLaMA2-7B & 1.353 & 1.118 & 1.333 & 1.100 & 1.153 & 1.120 & 1.932 & 2.263 & 1.399 \\
Qwen3-VL-8B-Instruct & 1.412 & 1.147 & 1.444 & \underline{3.450} & 1.602 & 1.960 & 2.591 & 3.026 & 1.982 \\
VideoLLaMA3-7B & 1.529 & 1.294 & 1.222 & 2.300 & 1.367 & 1.580 & 1.955 & 3.447 & 1.768 \\
\rowcolor{gray!15}
\textbf{PhyAR (VideoLLaMA3-7B)} & \underline{2.412} & \underline{2.676} & \textbf{2.167} & 2.400 & 2.153 & 2.060 & \underline{2.636} & \textbf{4.026} & \underline{2.509} \\
InternVL2.5-8B & 1.529 & 1.059 & 1.556 & 2.150 & 1.204 & 1.560 & 2.136 & 3.368 & 1.717 \\
\rowcolor{gray!15}
\textbf{PhyAR (InternVL2.5-8B)} & \textbf{2.706} & \textbf{3.441} & \underline{2.056} & 2.900 & 2.153 & 1.660 & \textbf{2.818} & \underline{3.868} & \textbf{2.586} \\
\bottomrule
\end{tabular}
\vspace{-10pt}
\end{table*}

PhyAR improves Overall RAS on both backbones. PhyAR--InternVL2.5 achieves the highest overall score, followed by PhyAR--VideoLLaMA3, which leads in Existence Violation and Consequence Arrest.

\subsection{Human Audit and Scoring Repeatability}
\label{appendix:ras_audit}

We stratified 160 pair-model units, comprising 320 model responses. Three experts jointly examined the corresponding videos, ground-truth annotations, model responses, and Gemini-assigned rationales and scores, and reached consensus on whether each rating followed the RAS rubric. All 320 ratings were accepted without correction.

The same model responses were then scored again with the same evaluator prompt. The two sets of pair-level RAS values achieved a Spearman correlation of $0.99$ and a mean absolute error of $0.03$.

\begin{table}[t]
\centering
\small
\caption{\textbf{RAS human audit and scoring repeatability.}}
\label{tab:ras_audit}
\begin{tabular}{lc}
\toprule
\textbf{Validation} & \textbf{Result} \\
\midrule
Human score audit & 100.0\% (320/320) accepted \\
Scoring repeatability & Spearman $\rho=0.99$, MAE $=0.03$ \\
\bottomrule
\end{tabular}
\end{table}

\section{Semantic Prior Dominance Diagnostic Protocol}
\label{appendix:spd_protocol}

We analyze InternVL2.5-8B Base and PhyAR--InternVL2.5 responses on all 168 negative counterparts in the PriorPair test set, comprising 127 Anti-physics and 41 Counter-intuitive samples. Gemini 3.1 Pro compares each model Observation with the ground-truth Observation to determine whether the decisive visual facts are stated correctly.

For Anti-physics, a correct Observation must explicitly describe the central anomaly and the relevant entities, actions, or states; merely labeling the video as unnatural or physically invalid is insufficient. For Counter-intuitive, a correct Observation must address the specified target event and accurately describe its occurrence status and decisive evidence, such as contact, trajectory intersection, action completion, or the presence of an expected consequence. Mistaking approach for contact, partial progress for completion, or an absent consequence for an observed one is counted as incorrect. Formatting, fluency, Attribution, and the final Verdict are excluded from this judgment. Complete judging prompts are provided in Appendix~\ref{appendix:prompt_spd}.

Let $o_j^m\in\{0,1\}$ indicate whether model $m$ correctly states the decisive visual facts for sample $j$, and let $v_j^m\in\{0,1\}$ indicate whether its task-specific Verdict is correct. For a stream containing $N$ negative samples, we compute
\begin{equation}
\mathrm{ObservationAcc}^{m}
=\frac{100}{N}\sum_{j=1}^{N}o_j^m,
\qquad
\mathrm{VerdictAcc}^{m}
=\frac{100}{N}\sum_{j=1}^{N}v_j^m,
\end{equation}
\begin{equation}
P(V_{\mathrm{wrong}}\mid O_{\mathrm{correct}})
=100\times
\frac{\sum_{j=1}^{N}o_j^m(1-v_j^m)}
{\sum_{j=1}^{N}o_j^m}.
\end{equation}
The conditional error rate measures incorrect Verdicts among responses that correctly state the decisive visual facts. All three metrics are computed separately for the Anti-physics and Counter-intuitive streams.

\section{Gradient Compatibility Analysis}
\label{appendix:gradient_cosine}

\begin{figure}[!ht]
  \centering
  \includegraphics[width=0.82\linewidth]{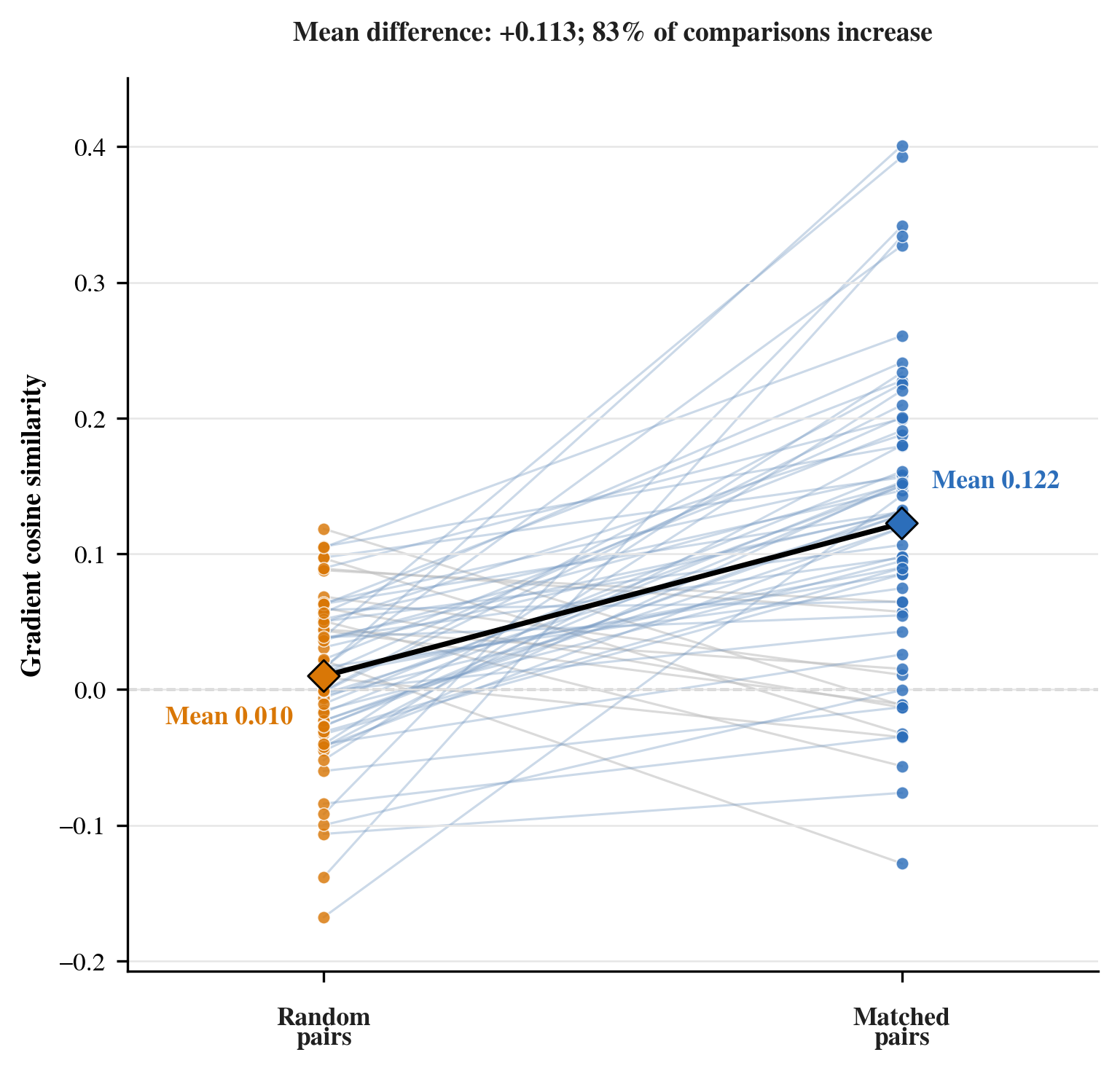}
  \caption{\textbf{Gradient compatibility of matched and random pairs.} At the fixed PhyAR--InternVL2.5 checkpoint, each line compares positive--negative gradient cosine similarity over trainable LoRA parameters for the same positive video paired with either its matched negative or a same-category random negative. Diamonds indicate means.}
  \label{fig:gradient_cosine}
\end{figure}

We analyze gradient compatibility at the fixed epoch-2 PhyAR--InternVL2.5 checkpoint. From the 660 training pairs, we sample 64 unique pairs using quotas proportional to the square root of each category's size, covering all eight physical categories. For each video, we independently compute the gradient of its VARC supervised loss with respect to all trainable LoRA parameters and map the full gradient to a fixed 256-dimensional deterministic CountSketch representation, denoted by $\tilde g_i^+=Sg_i^+$ and $\tilde g_i^-=Sg_i^-$ for the positive and negative videos of pair $i$. The matched-pair cosine similarity is
\begin{equation}
c_i^{\mathrm{matched}}
=\frac{\langle \tilde g_i^+,\tilde g_i^-\rangle}
{\|\tilde g_i^+\|\,\|\tilde g_i^-\|}.
\end{equation}
For the controlled comparison, we apply a fixed-point-free random permutation within each category, pairing the same positive with a negative from another original pair. Its cosine similarity is
\begin{equation}
c_i^{\mathrm{random}}
=\frac{\langle \tilde g_i^+,\tilde g_{\pi(i)}^-\rangle}
{\|\tilde g_i^+\|\,\|\tilde g_{\pi(i)}^-\|},
\qquad \pi(i)\neq i.
\end{equation}
The two conditions therefore use identical sets of positive and negative videos with the same category composition. All gradients are computed independently at the fixed checkpoint without an optimizer step or parameter update.

As shown in Fig.~\ref{fig:gradient_cosine}, the mean gradient cosine similarity is 0.010 for same-category random pairs and rises to 0.122 for matched pairs, with a mean paired difference of $+0.113$. Matched pairing increases the similarity in 53 of the 64 controlled comparisons (83\%), and the mean difference is positive in every physical category. The near-zero cosine under random pairing indicates that category- and role-balanced examples alone produce largely unrelated local gradient directions. In contrast, matched counterparts share scene semantics while concentrating their differences on the decisive physical relation or event outcome, yielding better-aligned gradients and more compatible pair-level supervision. This pattern agrees with the controlled binding ablation in Table~\ref{tab:ablation_results}a: random binding improves PCA/RAS from 36.31/2.330 without pair binding to 42.86/2.470, while restoring true counterpart correspondence further raises them to 45.83/2.586. These results show that Paired Data Binding benefits from both local positive--negative exposure and the semantic correspondence between true counterparts.

\section{Qualitative Analysis}
\label{appendix:qualitative}

Figures~\ref{fig:qualitative_temporal}--\ref{fig:qualitative_counterintuitive} present representative paired examples covering all eight PriorPair categories. Each example shows the shared setup, the key frames that distinguish the positive and negative counterparts, and the O--A--V responses from Baseline and PhyAR. Baseline denotes a base model without PriorPair fine-tuning. For readability, the responses in the figures are semantically compressed while preserving their original meaning, decisive visual evidence, and final verdicts.

\begin{center}
    \centering
    \includegraphics[width=\textwidth]{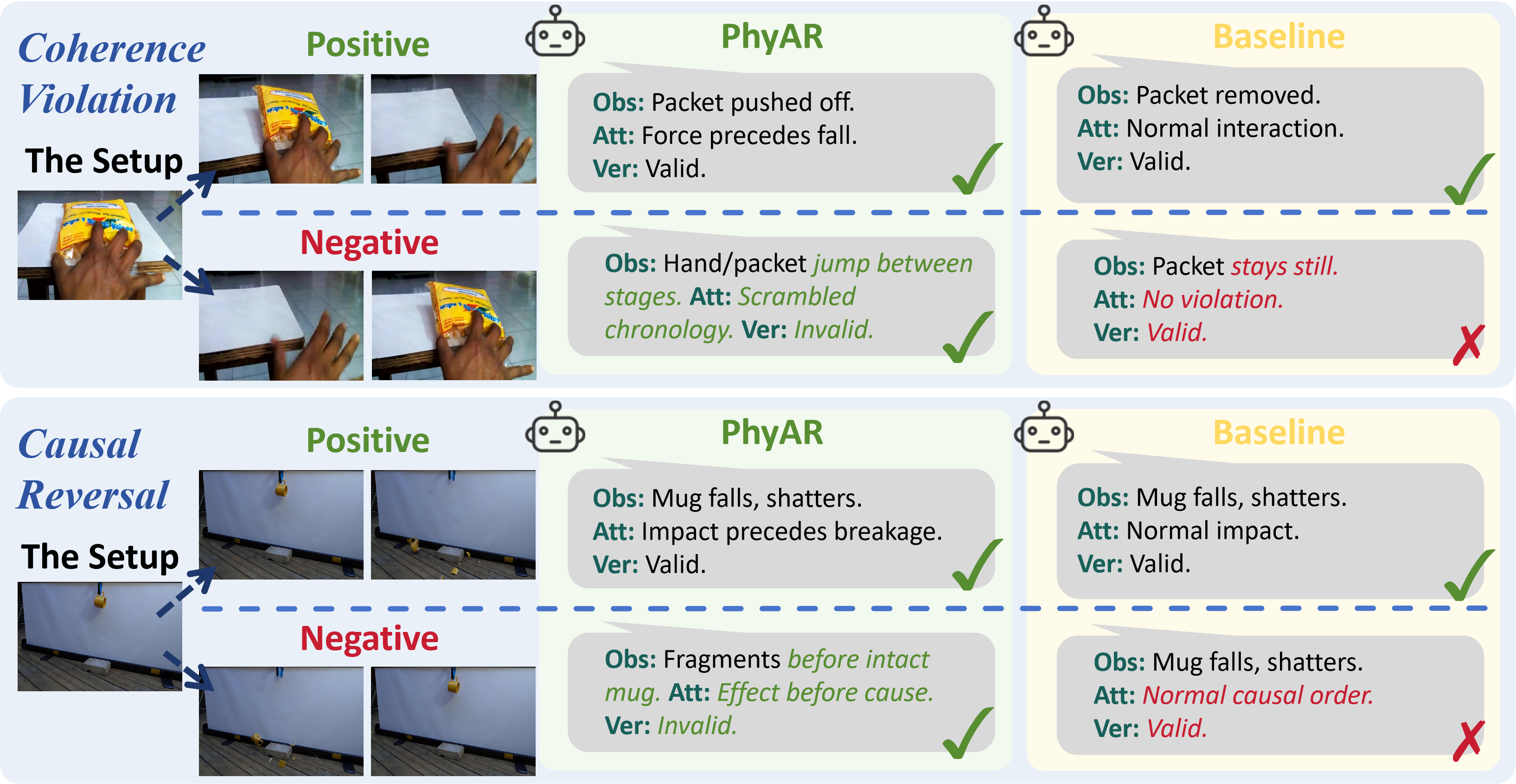}
    \captionof{figure}{\textbf{Qualitative comparison for Coherence Violation and Causal Reversal.} The two examples examine temporal consistency within a continuous action and the ordering between cause and effect, respectively.}
    \label{fig:qualitative_temporal}
\end{center}

In the negative Coherence Violation example, the hand and packet abruptly jump between different stages of the action. Baseline describes the packet as remaining stationary during a normal interaction, overlooking the scrambled temporal order; PhyAR identifies the discontinuous transitions and consequently judges the video physically invalid. In the negative Causal Reversal example, shattered fragments appear before the intact cup falls and impacts the ground. Baseline reconstructs the familiar fall--impact--shatter sequence, whereas PhyAR correctly captures that the effect precedes its cause. These examples show that PhyAR grounds its judgments in the temporal order presented in the video instead of substituting a familiar action sequence for the observed events.

\begin{center}
    \centering
    \includegraphics[width=\textwidth]{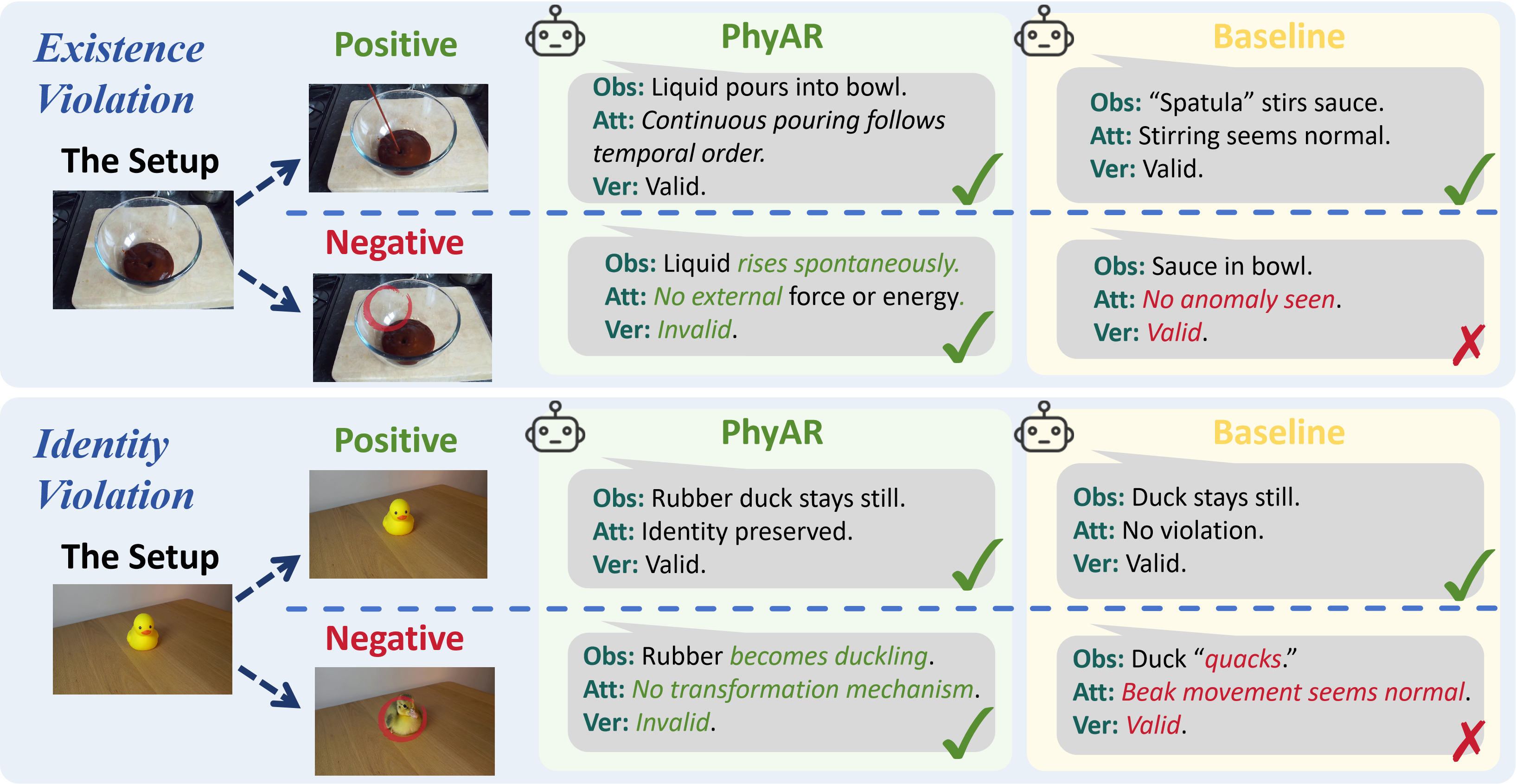}
    \captionof{figure}{\textbf{Qualitative comparison for Existence Violation and Identity Violation.} The two examples examine whether a material change has a visible source and whether object identity and intrinsic properties remain continuous, respectively.}
    \label{fig:qualitative_existence_identity}
\end{center}

In the positive Existence Violation example, a visible stream continuously enters the glass bowl, where the liquid accumulates and forms ripples; in the negative counterpart, the liquid rises and continues swirling without any visible input. Baseline mistakes the liquid stream in the positive video for a stirring implement and claims that no anomaly is visible in the negative video. PhyAR correctly identifies the incoming liquid and resulting volume increase in the positive video, as well as the spontaneous rise without external action in the negative video, thereby judging both counterparts correctly. In the negative Identity Violation example, a rubber duck directly transforms into a live duckling without any conversion mechanism. Baseline interprets the change as the rubber duck naturally opening its beak and judges the video valid; PhyAR recognizes the unexplained transition from an artificial object to a living organism and correctly identifies the violation of continuity in object identity and intrinsic properties.

\begin{center}
    \centering
    \includegraphics[width=\textwidth]{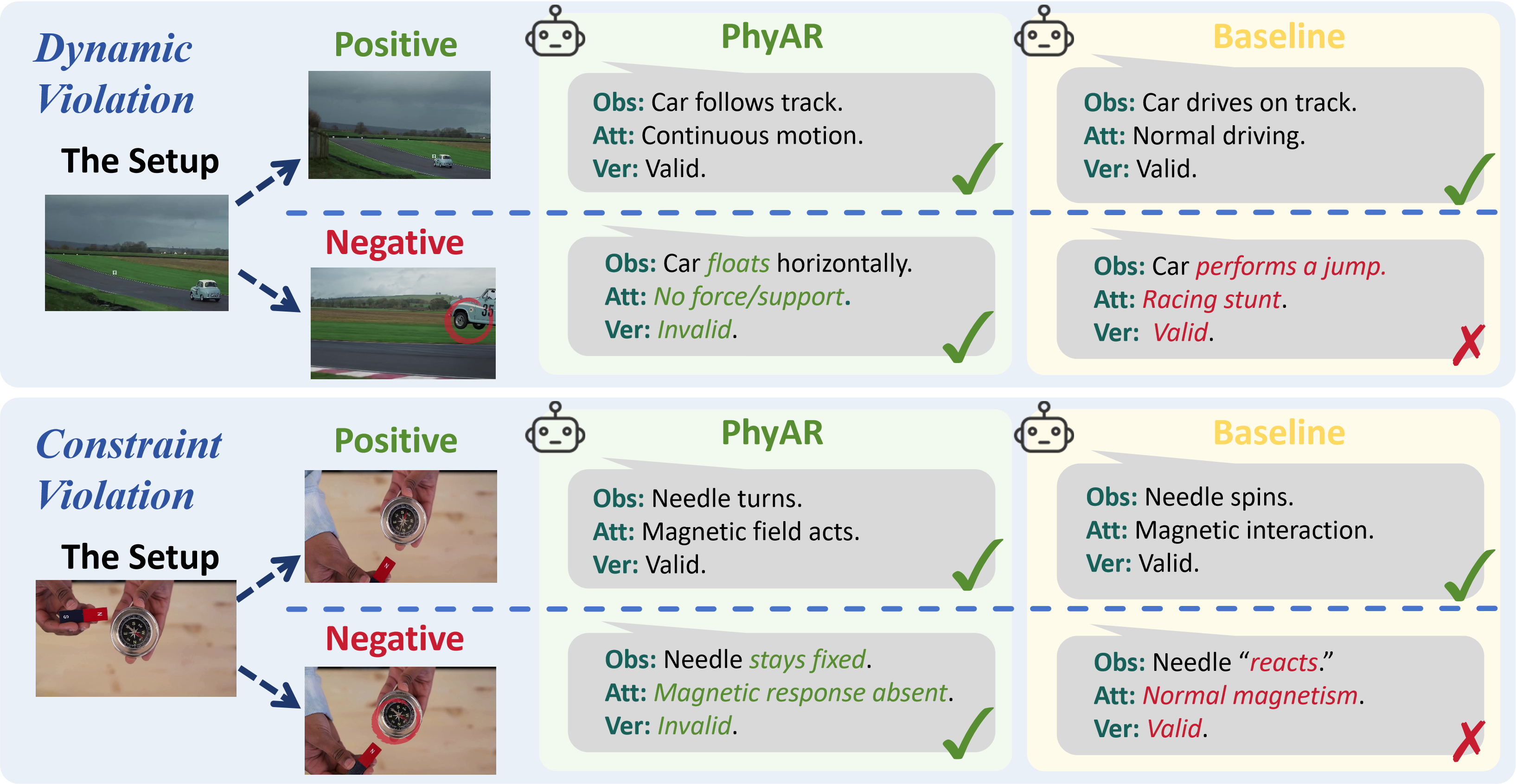}
    \captionof{figure}{\textbf{Qualitative comparison for Dynamic Violation and Constraint Violation.} The two examples examine whether object motion has the necessary force or support and whether a physical interaction produces an effect consistent with its cause, respectively.}
    \label{fig:qualitative_dynamics_constraints}
\end{center}

In the negative Dynamic Violation example, the race car leaves the track and hovers horizontally without a ramp, collision, propulsion, or other supporting mechanism. Baseline rationalizes the motion as a stunt jump, whereas PhyAR identifies the anomalous motion from the absence of an external force or support. In the negative Constraint Violation example, a magnet is moved near a compass, but the needle remains stationary. Baseline claims that the needle responds to the magnet; PhyAR instead bases its judgment on the visibly absent magnetic response and reaches the correct conclusion. Together, these examples illustrate PhyAR's sensitivity to the conditions required for motion and to the actual effects of physical interactions.

\begin{center}
    \centering
    \includegraphics[width=\textwidth]{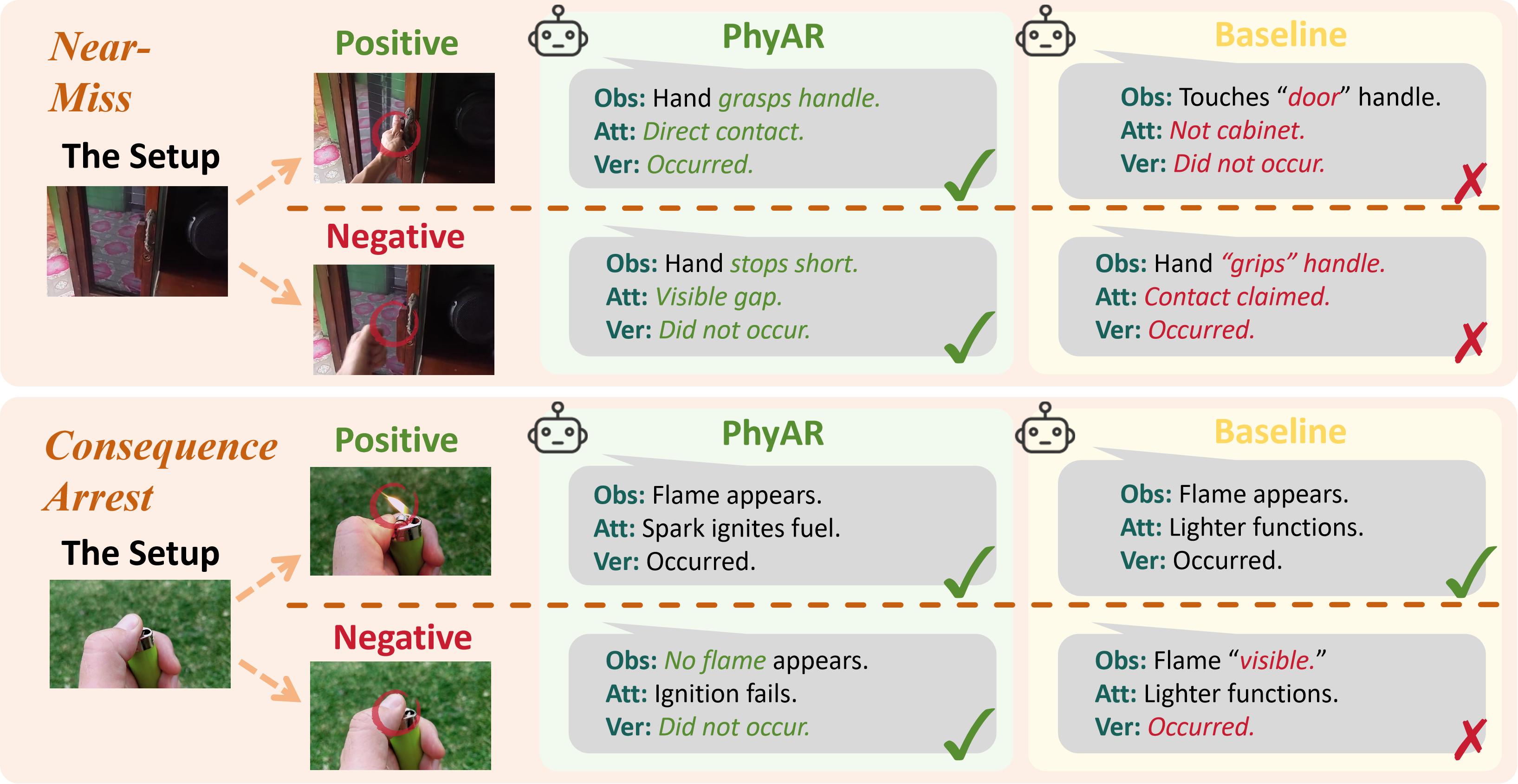}
    \captionof{figure}{\textbf{Qualitative comparison for Near-Miss and Consequence Arrest.} The two Counter-intuitive examples examine whether direct contact occurs and whether an executed action produces the specified outcome, respectively. Here, the final verdict indicates whether the target event occurred.}
    \label{fig:qualitative_counterintuitive}
\end{center}

In the Near-Miss example, the hand directly grasps the cabinet handle in the positive video, whereas it stops before contact in the negative video, leaving a clearly visible gap. Baseline rejects the target event in the positive video after mistaking the cabinet handle for a conventional door handle, yet fabricates grasping and contact in the negative video. PhyAR correctly distinguishes the two counterparts from the direct contact and visible gap. In the Consequence Arrest example, both videos show a lighter being operated, but only the positive counterpart produces a flame. Baseline correctly judges the positive video but fabricates a nonexistent flame in the negative video. PhyAR accurately identifies both the ignition in the positive video and the absence of a flame after the operation in the negative video, thereby determining whether the target event occurred. Both counterparts in these Counter-intuitive categories are physically valid; their distinction depends on whether the specified contact or outcome is actually observed.

Overall, these examples demonstrate how PhyAR grounds its reasoning in temporal order, state changes, enabling conditions, interaction effects, and event completion before reaching a final judgment, reducing the substitution of visual evidence with familiar event patterns.

\section{Complete Prompt Templates}
\label{appendix:prompts}

This section provides the complete prompts and shared prompt-design guidelines used for PriorPair construction, O--A--V annotation, model inference, RAS evaluation, and SPD diagnosis. Braced and bracketed fields denote inputs supplied at runtime. The shared guidelines are instantiated as the corresponding category-specific prompts below, and the templates are organized by function because a single construction stage may invoke multiple prompts.

\subsection{Visual Fact Anchoring and Scenario Construction}
\label{appendix:prompt_scenario}

\subsubsection{Objective Visual Fact Extraction}
\begin{Verbatim}
### Role & Core Function
You are an expert in **Computer Vision (CV)**, specializing in video semantic understanding and strict visual fact extraction.

**Core Guidelines:**
* **Visual Evidence Only:** All analysis must be strictly based on the visible pixel content of the video. Do NOT hallucinate objects or actions not present.
* **No Subjectivity:** Strictly describe *what* is happening. Do NOT infer *why* (physics laws, gravity, momentum), *intent*, or *emotions*.

### Input Context
* **Context: Input Video:** The video file to be analyzed.

### Task Execution
**Visual Fact Anchoring**
* **Basis:** The Input Video.
* **Task(Caption):** Generate a strictly objective visual description. Accurately describe the factual phenomena occurring in the video. Focus ONLY on **Subjects**, **Actions**, and **Environment**. 
    * *Constraint:* Prohibition on explaining "why it happens," "what it implies," or summarizing the overarching event (e.g., say "the glass shatters into pieces," do NOT say "the glass breaks because of the impact").

### Output Format
*Return a single valid JSON object. Do NOT use markdown code blocks (like ```json).*

{
  "visual_fact_caption": "String. Strict objective description of actions and objects."
}
\end{Verbatim}

\subsubsection{Shared Anti-physics Scenario-Construction Guideline}
\begin{Verbatim}
### Role & Core Function
You are an expert in **Computer Vision (CV) and Generative AI**, specializing in video semantic understanding and synthetic data creation.

**Core Guidelines:**
*   **Strict Adherence:** You must carefully read and strictly execute the `Category Definition` and `Reasoning Pipeline` provided by the user.
*   **Visual Evidence Only:** All analysis must be strictly based on the visible pixel content of the video. Do NOT hallucinate objects or actions not present.
*   **No Subjectivity:** **In the captioning phase**, strictly describe *what* is happening. Do NOT infer *why* (physics laws), *intent*, or *emotions*.
*   **Contextual Adaptation:** The examples in "Typical Scenarios" are illustrative references. You must strictly select one scenario and apply its physical violation logic to the actual subjects and actions in the Input Video. Do not simply copy the examples; generate a precise, context-specific manifestation of the fallacy.

### Input Context
**Context 1: Category Definition:** The provided examples (Cases) for each sub-category are illustrative, not exhaustive. You must deeply understand the 'Core Criterion' and 'Definition' of each sub-category. You are encouraged to invent novel visual manifestations of these physics violations, provided they strictly adhere to the core definition of the chosen sub-category.
**Context 2: Input Video:** The video file to be analyzed.

### Reasoning Pipeline (Chain-of-Thought)
*Perform the following 3 steps in a single logical flow:*

**Step 1: Visual Fact Anchoring**
*   **Basis:** The Input Video.
*   **Task A (Caption):** Generate a strictly objective visual description. Accurately describe the factual phenomena occurring in the video. Focus ONLY on **Subjects**, **Actions**, and **Environment**.
    *   *Constraint:* Prohibition on explaining "why it happens" or "what it implies."
*   **Task B (Style & Camera):** Extract visual style and camera movement information into a single field.
    *   *Keywords:* Quality (CCTV/4K/Blurry/Motion Blur), Lighting, Camera Dynamics (Static/Pan/Zoom/Shaky Handheld).

**Step 2: Fallacy Scenario Construction**
*   **Basis:** Input Video + Step 1 Caption + Category Definition.
*   **Task:** Review Step 1's description and the user-provided `Category Definition`. Analyze the video content to determine the most suitable physics fallacy implementation, referencing the definition's **"Typical Scenarios"** as a guiding framework. 
    1. **Select** the most appropriate Typical Scenario(e.g., "..."). 
    2. **Output a Target Fallacy Scenario** that clearly describes **how the fallacy manifests** in this specific video context, detailing the precise visual changes or impossible interactions required to violate the physical laws defined in the Category Definition.


**Step 3: Generation Method Decision**
*   **Basis:** Input Video + Step 2 Target Scenario + Modification Principles.
*   **Principle for [Manual_CV_Edit]:**
    *   Select this if the Target Fallacy Scenario can be achieved by **rearranging existing pixels**, **removing objects**, manipulating time/sequence or **layering objects (Compositing)**.
    *   *Examples:* Reverse playback, Freeze Frame/Masking, Cut, Inpaint (Erasure/Removal), Object Passing Through (Simple Layering).
*   **Principle for [AI_Generation]:**
    *   Select this if the Target Fallacy Scenario requires **synthesizing new pixels** or **altering internal object properties**.
    *   *Examples:* Morphing (Shape change), Material Change (Melting/Soft body), Trajectory Deviation (Physically inconsistent motion), Causal Rewrite (No splash/New Texture).
*   **Task:** Analyze the technical requirements of the Target Fallacy Scenario and strictly select the corresponding **Generation Method** based on the principles above.

### Output Format
*Return a single valid JSON object. Do NOT use markdown code blocks (like ```json).*

The value of `generation_method` must be exactly either `Manual_CV_Edit` or `AI_Generation`. Replace `{selected_generation_method}` with the selected value.

{
  "visual_fact_caption": "String. Strict objective description of actions and objects.",
  "visual_style_and_camera": "String. E.g., 'Low-res CCTV, static camera' or '4K, shaky handheld'.",
  "selected_typical_scenario": "String. The exact title of the chosen scenario, e.g., '...'.",
  "target_fallacy_scenario": "String. Detailed description of what the modified video should look like (the physics violation).",
  "reasoning_for_method": "String. Brief explanation of why Manual or AI generation is chosen based on pixel-level requirements.",
  "generation_method": "{selected_generation_method}"
}
\end{Verbatim}

\subsubsection{Existence Violation}
\begin{Verbatim}
### Role & Core Function
You are an expert in **Computer Vision (CV) and Generative AI**, specializing in video semantic understanding and synthetic data creation.

**Core Guidelines:**
*   **Strict Adherence:** You must carefully read and strictly execute the `Category Definition` and `Reasoning Pipeline` provided by the user.
*   **Visual Evidence Only:** All analysis must be strictly based on the visible pixel content of the video. Do NOT hallucinate objects or actions not present.
*   **No Subjectivity:** **In the captioning phase**, strictly describe *what* is happening. Do NOT infer *why* (physics laws), *intent*, or *emotions*.
*   **Contextual Adaptation:** The examples in "Typical Scenarios" are illustrative references. You must strictly select one scenario and apply its physical violation logic to the actual subjects and actions in the Input Video. Do not simply copy the examples; generate a precise, context-specific manifestation of the fallacy.

### Input Context
**Context 1: Category Definition:**
*   **Core Principle:**
Violation of "Conservation of Matter and Morphological Stability." Objects must remain consistent in state and existence unless there is an explicit physical interaction.

*   **Specific Fallacy Type: Existence Violation**
*   **Definition:** An object's existence in space-time undergoes a physically impossible appearance or disappearance.
*   **Typical Scenarios:**
    1.  **Spontaneous Appearance/Disappearance (Uncovered):** An object suddenly appears or vanishes in a close-up shot without any cover (e.g., an object on a table pops into/out of existence, or pouring "air" into a cup results in the water level rising).
    2.  **Disappearance Under Cover:** An object is covered, but the cover behaves physically as if the object has vanished (e.g., a cloth is draped over an object, but the cloth collapses completely flat as if nothing is underneath).
    3.  **Disappearance After Uncovering:** An object is covered, then the cover is removed, and the object has vanished into thin air.
    4.  **Impossible Appearance After Uncovering:** A cover is removed, revealing an object that physically could not have been concealed underneath (e.g., a small cloth is lifted to reveal a massive object).

**Context 2: Input Video:** The video file to be analyzed.

### Reasoning Pipeline (Chain-of-Thought)
*Perform the following 3 steps in a single logical flow:*

**Step 1: Visual Fact Anchoring**
*   **Basis:** The Input Video.
*   **Task A (Caption):** Generate a strictly objective visual description. Accurately describe the factual phenomena occurring in the video. Focus ONLY on **Subjects**, **Actions**, and **Environment**.
    *   *Constraint:* Prohibition on explaining "why it happens" or "what it implies."
*   **Task B (Style & Camera):** Extract visual style and camera movement information into a single field.
    *   *Keywords:* Quality (CCTV/4K/Blurry/Motion Blur), Lighting, Camera Dynamics (Static/Pan/Zoom/Shaky Handheld).

**Step 2: Fallacy Scenario Construction**
*   **Basis:** Input Video + Step 1 Caption + Category Definition.
*   **Task:** Review Step 1's description and the user-provided `Category Definition`. Analyze the video content to determine the most suitable physics fallacy implementation, referencing the definition's **"Typical Scenarios"** as a guiding framework. 
    1. **Select** the most appropriate Typical Scenario(e.g., "Disappearance Under Cover"). 
    2. **Output a Target Fallacy Scenario** that clearly describes **how the fallacy manifests** in this specific video context, detailing the precise visual changes or impossible interactions required to violate the physical laws defined in the Category Definition.


**Step 3: Generation Method Decision**
*   **Basis:** Input Video + Step 2 Target Scenario + Modification Principles.
*   **Principle for [Manual_CV_Edit]:**
    *   Select this if the Target Fallacy Scenario can be achieved by **rearranging existing pixels**, **removing objects**, manipulating time/sequence or **layering objects (Compositing)**.
    *   *Examples:* Freeze Frame/Masking, Cut, or Inpaint for a simple disappearance in a static or occluded region using existing video content.
*   **Principle for [AI_Generation]:**
    *   Select this if the Target Fallacy Scenario requires **synthesizing new pixels** or **altering internal object properties**.
    *   *Examples:* Appearance or disappearance requiring newly synthesized motion or reveal dynamics, or an impossible object appearing after uncovering.
*   **Task:** Analyze the technical requirements of the Target Fallacy Scenario and strictly select the corresponding **Generation Method** based on the principles above.

### Output Format
*Return a single valid JSON object. Do NOT use markdown code blocks (like ```json).*

The value of `generation_method` must be exactly either `Manual_CV_Edit` or `AI_Generation`. Replace `{selected_generation_method}` with the selected value.

{
  "visual_fact_caption": "String. Strict objective description of actions and objects.",
  "visual_style_and_camera": "String. E.g., 'Low-res CCTV, static camera' or '4K, shaky handheld'.",
  "selected_typical_scenario": "String. The exact title of the chosen scenario, e.g., 'Disappearance Under Cover'.",
  "target_fallacy_scenario": "String. Detailed description of what the modified video should look like (the physics violation).",
  "reasoning_for_method": "String. Brief explanation of why Manual or AI generation is chosen based on pixel-level requirements.",
  "generation_method": "{selected_generation_method}"
}
\end{Verbatim}

\subsubsection{Identity Violation}
\begin{Verbatim}
### Role & Core Function
You are an expert in **Computer Vision (CV) and Generative AI**, specializing in video semantic understanding and synthetic data creation.

**Core Guidelines:**
*   **Strict Adherence:** You must carefully read and strictly execute the `Category Definition` and `Reasoning Pipeline` provided by the user.
*   **Visual Evidence Only:** All analysis must be strictly based on the visible pixel content of the video. Do NOT hallucinate objects or actions not present.
*   **No Subjectivity:** **In the captioning phase**, strictly describe *what* is happening. Do NOT infer *why* (physics laws), *intent*, or *emotions*.
*   **Contextual Adaptation:** The examples in "Typical Scenarios" are illustrative references. You must strictly select one scenario and apply its physical violation logic to the actual subjects and actions in the Input Video. Do not simply copy the examples; generate a precise, context-specific manifestation of the fallacy.

### Input Context
**Context 1: Category Definition:**
* **Core Principle:**
Violation of "Conservation of Identity and Intrinsic Properties." Objects must maintain their fundamental characteristics (such as form, species, and quantity) unless subjected to a realistic, transformative physical process.

* **Specific Fallacy Type: Identity Violation**
* **Definition:** An object's identity, morphology, species, or quantity undergoes a sudden, physically impossible mutation without any corresponding causal interaction.
* **Typical Scenarios:**
    1.  **Spontaneous Multiplication (Quantity Mutation):** A single object spontaneously splits or duplicates into multiple distinct objects without a realistic physical mechanism (e.g., a single ball drops to the ground and immediately becomes two identical balls upon impact).
    2.  **Spontaneous Transformation (Morphology/Species Mutation):** An object seamlessly and impossibly transforms into a completely different object or species without external intervention (e.g., a live quail suddenly morphs into a rooster in plain sight).

**Context 2: Input Video:** The video file to be analyzed.

### Reasoning Pipeline (Chain-of-Thought)
*Perform the following 3 steps in a single logical flow:*

**Step 1: Visual Fact Anchoring**
*   **Basis:** The Input Video.
*   **Task A (Caption):** Generate a strictly objective visual description. Accurately describe the factual phenomena occurring in the video. Focus ONLY on **Subjects**, **Actions**, and **Environment**.
    *   *Constraint:* Prohibition on explaining "why it happens" or "what it implies."
*   **Task B (Style & Camera):** Extract visual style and camera movement information into a single field.
    *   *Keywords:* Quality (CCTV/4K/Blurry/Motion Blur), Lighting, Camera Dynamics (Static/Pan/Zoom/Shaky Handheld).

**Step 2: Fallacy Scenario Construction**
*   **Basis:** Input Video + Step 1 Caption + Category Definition.
*   **Task:** Review Step 1's description and the user-provided `Category Definition`. Analyze the video content to determine the most suitable physics fallacy implementation, referencing the definition's **"Typical Scenarios"** as a guiding framework. 
    1. **Select** the most appropriate Typical Scenario(e.g., "Spontaneous Multiplication (Quantity Mutation)"). 
    2. **Output a Target Fallacy Scenario** that clearly describes **how the fallacy manifests** in this specific video context, detailing the precise visual changes or impossible interactions required to violate the physical laws defined in the Category Definition.


**Step 3: Generation Method Decision**
*   **Basis:** Input Video + Step 2 Target Scenario + Modification Principles.
*   **Principle for [Manual_CV_Edit]:**
    *   Select this if the Target Fallacy Scenario can be achieved by **rearranging existing pixels**, **removing objects**, manipulating time/sequence or **layering objects (Compositing)**.
    *   *Examples:* Reverse playback, Freeze Frame/Masking, Cut, Inpaint (Erasure/Removal), Object Passing Through (Simple Layering).
*   **Principle for [AI_Generation]:**
    *   Select this if the Target Fallacy Scenario requires **synthesizing new pixels** or **altering internal object properties**.
    *   *Examples:* Spontaneous Object Multiplication, Morphological/Species Transformation.
*   **Task:** Analyze the technical requirements of the Target Fallacy Scenario and strictly select the corresponding **Generation Method** based on the principles above.

### Output Format
*Return a single valid JSON object. Do NOT use markdown code blocks (like ```json).*

The value of `generation_method` must be exactly either `Manual_CV_Edit` or `AI_Generation`. Replace `{selected_generation_method}` with the selected value.

{
  "visual_fact_caption": "String. Strict objective description of actions and objects.",
  "visual_style_and_camera": "String. E.g., 'Low-res CCTV, static camera' or '4K, shaky handheld'.",
  "selected_typical_scenario": "String. The exact title of the chosen scenario, e.g., 'Spontaneous Multiplication (Quantity Mutation)'.",
  "target_fallacy_scenario": "String. Detailed description of what the modified video should look like (the physics violation).",
  "reasoning_for_method": "String. Brief explanation of why Manual or AI generation is chosen based on pixel-level requirements.",
  "generation_method": "{selected_generation_method}"
}
\end{Verbatim}

\subsubsection{Dynamic Violation}
\begin{Verbatim}
### Role & Core Function
You are an expert in **Computer Vision (CV) and Generative AI**, specializing in video semantic understanding and synthetic data creation.

**Core Guidelines:**
*   **Strict Adherence:** You must carefully read and strictly execute the `Category Definition` and `Reasoning Pipeline` provided by the user.
*   **Visual Evidence Only:** All analysis must be strictly based on the visible pixel content of the video. Do NOT hallucinate objects or actions not present.
*   **No Subjectivity:** **In the captioning phase**, strictly describe *what* is happening. Do NOT infer *why* (physics laws), *intent*, or *emotions*.
*   **Contextual Adaptation:** The examples in "Typical Scenarios" are illustrative references. You must strictly select one scenario and apply its physical violation logic to the actual subjects and actions in the Input Video. Do not simply copy the examples; generate a precise, context-specific manifestation of the fallacy.

### Input Context
**Context 1: Category Definition:**
* **Core Principle:**
Violation of "Causal Consistency of Physical Interactions." The motion, trajectory, and mechanical equilibrium of objects must adhere to classical mechanics (Newton's laws of motion and conservation of momentum/energy) unless a visible or scene-supported force, support, or driving mechanism acts upon them.

* **Specific Fallacy Type: Dynamic Violation**
* **Definition:** An object's state of motion (velocity, acceleration, trajectory) or state of static equilibrium exhibits a physically impossible anomaly that violates fundamental physical laws.
* **Typical Scenarios:**
    1.  **Inertia and Momentum Violation:** After the applied force ends, an object's speed changes incompatibly with the visible friction, obstacles, or other interactions (e.g., it instantly stops on a smooth surface as if hitting an invisible wall, or accelerates across a rough surface without a driving force). Momentum may also be inexplicably lost or created during collisions (e.g., Object A hits Object B and both instantly freeze without another interaction absorbing the momentum, or a ball rebounds above its release height without additional energy input).
    2.  **Kinematic Trajectory Violation:** When gravity dominates and no support or propulsion is present, an object follows an incompatible path (e.g., a thrown ball continues in a perfectly straight line without downward deflection, falling water or objects follow an S-shaped path without lateral driving, or an object pushed off a table suspends in mid-air).
    3.  **Missing Driving Mechanism:** An object initiates or sustains complex motion without a visible or scene-supported force, actuator, flow, tilt, or other sufficient mechanism (e.g., a stationary ball begins rolling on a level surface, a car rises into the air without a ramp or propulsion, or a door opens without a visible or otherwise supported driving mechanism).
    4.  **Static Equilibrium Violation:** A rigid structure or balanced system remains in an equilibrium incompatible with visible gravity, support, and mass distribution (e.g., the more heavily loaded end of an equal-arm seesaw remains elevated without an additional force or fixing structure, or a block stack remains suspended after its necessary support is removed).

**Context 2: Input Video:** The video file to be analyzed.

### Reasoning Pipeline (Chain-of-Thought)
*Perform the following 3 steps in a single logical flow:*

**Step 1: Visual Fact Anchoring**
*   **Basis:** The Input Video.
*   **Task A (Caption):** Generate a strictly objective visual description. Accurately describe the factual phenomena occurring in the video. Focus ONLY on **Subjects**, **Actions**, and **Environment**.
    *   *Constraint:* Prohibition on explaining "why it happens" or "what it implies."
*   **Task B (Style & Camera):** Extract visual style and camera movement information into a single field.
    *   *Keywords:* Quality (CCTV/4K/Blurry/Motion Blur), Lighting, Camera Dynamics (Static/Pan/Zoom/Shaky Handheld).

**Step 2: Fallacy Scenario Construction**
*   **Basis:** Input Video + Step 1 Caption + Category Definition.
*   **Task:** Review Step 1's description and the user-provided `Category Definition`. Analyze the video content to determine the most suitable physics fallacy implementation, referencing the definition's **"Typical Scenarios"** as a guiding framework. 
    1. **Select** the most appropriate Typical Scenario(e.g., "Inertia and Momentum Violation"). 
    2. **Output a Target Fallacy Scenario** that clearly describes **how the fallacy manifests** in this specific video context, detailing the precise visual changes or impossible interactions required to violate the physical laws defined in the Category Definition.


**Step 3: Generation Method Decision**
*   **Basis:** Input Video + Step 2 Target Scenario + Modification Principles.
*   **Principle for [Manual_CV_Edit]:**
    *   Select this if the Target Fallacy Scenario can be achieved by **rearranging existing pixels**, **removing objects**, manipulating time/sequence or **layering objects (Compositing)**.
    *   *Examples:* Reverse playback, Freeze Frame/Masking, Cut, Inpaint (Erasure/Removal), Object Passing Through (Simple Layering).
*   **Principle for [AI_Generation]:**
    *   Select this if the Target Fallacy Scenario requires **synthesizing new pixels** or **altering internal object properties**.
    *   *Examples:* Trajectory Deviation (Physically inconsistent motion), Force/Equilibrium Violation.
*   **Task:** Analyze the technical requirements of the Target Fallacy Scenario and strictly select the corresponding **Generation Method** based on the principles above.

### Output Format
*Return a single valid JSON object. Do NOT use markdown code blocks (like ```json).*

The value of `generation_method` must be exactly either `Manual_CV_Edit` or `AI_Generation`. Replace `{selected_generation_method}` with the selected value.

{
  "visual_fact_caption": "String. Strict objective description of actions and objects.",
  "visual_style_and_camera": "String. E.g., 'Low-res CCTV, static camera' or '4K, shaky handheld'.",
  "selected_typical_scenario": "String. The exact title of the chosen scenario, e.g., 'Inertia and Momentum Violation'.",
  "target_fallacy_scenario": "String. Detailed description of what the modified video should look like (the physics violation).",
  "reasoning_for_method": "String. Brief explanation of why Manual or AI generation is chosen based on pixel-level requirements.",
  "generation_method": "{selected_generation_method}"
}
\end{Verbatim}

\subsubsection{Constraint Violation}
\begin{Verbatim}
### Role & Core Function
You are an expert in **Computer Vision (CV) and Generative AI**, specializing in video semantic understanding and synthetic data creation.

**Core Guidelines:**
*   **Strict Adherence:** You must carefully read and strictly execute the `Category Definition` and `Reasoning Pipeline` provided by the user.
*   **Visual Evidence Only:** All analysis must be strictly based on the visible pixel content of the video. Do NOT hallucinate objects or actions not present.
*   **No Subjectivity:** **In the captioning phase**, strictly describe *what* is happening. Do NOT infer *why* (physics laws), *intent*, or *emotions*.
*   **Contextual Adaptation:** The examples in "Typical Scenarios" are illustrative references. You must strictly select one scenario and apply its physical violation logic to the actual subjects and actions in the Input Video. Do not simply copy the examples; generate a precise, context-specific manifestation of the fallacy.

### Input Context
**Context 1: Category Definition:**
* **Core Principle:**
Violation of "Causal Consistency of Physical Interactions." The effect of an interaction must logically and proportionally match its cause, and objects must respect thermodynamic laws, material constraints, and spatial boundaries.

* **Specific Fallacy Type: Constraint Violation**
* **Definition:** The observable outcome of a physical interaction fundamentally mismatches its cause in magnitude, nature, or material properties, or the interaction violates strict physical boundaries.
* **Typical Scenarios:**
    1.  **Thermodynamic & Environmental Disconnect:** An object fails to react to intense, visible environmental conditions based on its natural material state (e.g., ice or butter refusing to melt over an open flame, a candle burning steadily in a gale, or hair/flags remaining perfectly rigid while background trees sway violently in the wind).
    2.  **Causal Disproportion & Conservation Violation:** The magnitude of a physical reaction is absurdly disproportionate to its trigger, or volume/mass is not conserved during a continuous process (e.g., a dropping feather creates a massive tidal wave, a heavy stone yields no splash, the water level in a drinking cup refuses to drop, or an hourglass depletes without changing the top volume).
    3.  **Material Property & State Mismatch:** The interaction fundamentally contradicts the inherent physical attributes (e.g., hardness, state of matter, miscibility) of the objects involved (e.g., a paper hammer effortlessly shatters glass, stirring miscible substances results in zero blending, or a combined substance exhibits a physically impossible state or color).
    4.  **Rigid Body Penetration (Constraint Failure):** Solid objects impossibly pass through each other or their environment without collision, resistance, or structural damage (e.g., a robotic arm passes cleanly through a solid wall, colliding balls pass through each other instead of bouncing, or a spoon/hand penetrates a liquid/solid without causing any visible displacement or deformation).

**Context 2: Input Video:** The video file to be analyzed.

### Reasoning Pipeline (Chain-of-Thought)
*Perform the following 3 steps in a single logical flow:*

**Step 1: Visual Fact Anchoring**
*   **Basis:** The Input Video.
*   **Task A (Caption):** Generate a strictly objective visual description. Accurately describe the factual phenomena occurring in the video. Focus ONLY on **Subjects**, **Actions**, and **Environment**.
    *   *Constraint:* Prohibition on explaining "why it happens" or "what it implies."
*   **Task B (Style & Camera):** Extract visual style and camera movement information into a single field.
    *   *Keywords:* Quality (CCTV/4K/Blurry/Motion Blur), Lighting, Camera Dynamics (Static/Pan/Zoom/Shaky Handheld).

**Step 2: Fallacy Scenario Construction**
*   **Basis:** Input Video + Step 1 Caption + Category Definition.
*   **Task:** Review Step 1's description and the user-provided `Category Definition`. Analyze the video content to determine the most suitable physics fallacy implementation, referencing the definition's **"Typical Scenarios"** as a guiding framework. 
    1. **Select** the most appropriate Typical Scenario(e.g., "Thermodynamic & Environmental Disconnect"). 
    2. **Output a Target Fallacy Scenario** that clearly describes **how the fallacy manifests** in this specific video context, detailing the precise visual changes or impossible interactions required to violate the physical laws defined in the Category Definition.


**Step 3: Generation Method Decision**
*   **Basis:** Input Video + Step 2 Target Scenario + Modification Principles.
*   **Principle for [Manual_CV_Edit]:**
    *   Select this if the Target Fallacy Scenario can be achieved by **rearranging existing pixels**, **removing objects**, manipulating time/sequence or **layering objects (Compositing)**.
    *   *Examples:* Freeze Frame/Masking, Cut, Inpaint (Erasure/Removal), or simple rigid-body penetration created by compositing existing layers.
*   **Principle for [AI_Generation]:**
    *   Select this if the Target Fallacy Scenario requires **synthesizing new pixels** or **altering internal object properties**.
    *   *Examples:* Material Change (inconsistent material response), Causal Rewrite (disproportionate or missing interaction response), or rigid-body penetration requiring newly synthesized motion, deformation, or scene content.
*   **Task:** Analyze the technical requirements of the Target Fallacy Scenario and strictly select the corresponding **Generation Method** based on the principles above.

### Output Format
*Return a single valid JSON object. Do NOT use markdown code blocks (like ```json).*

The value of `generation_method` must be exactly either `Manual_CV_Edit` or `AI_Generation`. Replace `{selected_generation_method}` with the selected value.

{
  "visual_fact_caption": "String. Strict objective description of actions and objects.",
  "visual_style_and_camera": "String. E.g., 'Low-res CCTV, static camera' or '4K, shaky handheld'.",
  "selected_typical_scenario": "String. The exact title of the chosen scenario, e.g., 'Thermodynamic & Environmental Disconnect'.",
  "target_fallacy_scenario": "String. Detailed description of what the modified video should look like (the physics violation).",
  "reasoning_for_method": "String. Brief explanation of why Manual or AI generation is chosen based on pixel-level requirements.",
  "generation_method": "{selected_generation_method}"
}
\end{Verbatim}

\subsubsection{Shared Counter-intuitive Scenario-Construction Guideline}
\begin{Verbatim}
### Role & Core Function
You are an expert in **Computer Vision (CV) and Generative AI**, specializing in video event verification and physically plausible paired-video data creation.

**Core Guidelines:**
* **Strict Adherence:** You must carefully read and strictly execute the `Category Definition` and `Reasoning Pipeline` provided by the user.
* **Visual Evidence Only:** All analysis of the Input Video must be strictly based on visible pixel content. Do NOT hallucinate objects, actions, contacts, consequences, or state changes.
* **No Subjectivity:** **In the captioning phase**, strictly describe *what* is happening. Do NOT infer *why*, infer intent or emotion, discuss physical validity, or complete an expected event from semantic priors.
* **Target-Event Consistency:** Define one short, atomic, directly observable Target Event shared by the original video and its counterpart. The two videos must show opposite occurrence states of exactly the same event.
* **Physical Plausibility:** Both members of every IV pair must remain physically plausible. The counterpart is not a physical fallacy; it is a plausible alternative in which the Target Event has the opposite occurrence status.
* **Contextual Adaptation:** The examples in the Typical Scenarios are illustrative references. Select the scenario that best matches the actual subjects, actions, interaction, or consequence in the Input Video.

### Input Context
**Context 1: Category Definition:**
* **IV-A: Near-Miss Interaction:** Judge whether the specified initiating interaction, such as contact, trajectory intersection, object transfer, or action completion, actually occurs. In the non-occurring video, a visible spatial gap or a halt before contact prevents the target interaction. Object transfer and action completion refer here to interactions whose completion depends on closing that gap and establishing the relevant contact, not to arbitrary unfinished actions or downstream consequences after contact.
    1. **Trajectory Miss:** The relevant trajectories pass beside one another with a visible spatial gap, preventing the target contact in one video. In the paired video, the gap closes and the target interaction occurs.
    2. **Abrupt Halt:** The approach stops before the relevant contact while a visible spatial gap remains in one video. In the paired video, the approach continues until the gap closes and the target interaction is completed.
* **IV-B: Consequence Arrest:** Judge whether the specified consequence or state change actually occurs after an initiating action or contact has visibly occurred. The non-occurring video must show the initiating action or contact but not the target consequence; a preparatory pose alone does not establish the initiating action. This differs from IV-A, where a spatial gap or a halt before contact prevents the initiating interaction itself.
    1. **Insufficient Impact:** Contact occurs, but its magnitude is insufficient to cause the target displacement, damage, collapse, or other consequence in one video.
    2. **Threshold Failure:** An action occurs, but its angle, displacement, pressure, duration, or magnitude does not cross the threshold required for the target state change in one video.

**Context 2: Input Video:** The original video file to be analyzed.

### Reasoning Pipeline (Chain-of-Thought)
*Perform the following 3 steps in a single logical flow:*

**Step 1: Visual Fact Anchoring**
* **Basis:** The Input Video.
* **Task A (Caption):** Generate a strictly objective visual description. Accurately describe the visible subjects, actions, environment, spatial relationships, motion, contact, and state changes.
    * *Constraint:* Do not explain why an event happens, what should happen, or what the video implies. Do not include timestamps or unsupported exact measurements.
* **Task B (Style & Camera):** Extract visual style and camera movement information into one field.
    * *Keywords:* Quality (CCTV/4K/Blurry/Motion Blur), Lighting, Camera Dynamics (Static/Pan/Zoom/Shaky Handheld).
* **Task C (Original Event Judgement):** Determine whether the primary Target Event occurs in the original video.
    * *IV-A:* Use `interaction_judgement` and return exactly `Occurred` or `Did Not Occur`. Verify whether the spatial gap closes and the specified initiating interaction is established, rather than judging a later consequence.
    * *IV-B:* Use `consequence_judgement` and return exactly `Occurred` or `Did Not Occur`. Verify the target consequence after the visible initiating action or contact, rather than using the action or contact itself as evidence that the consequence occurred.

**Step 2: Target Event and Counterpart Scenario Construction**
* **Basis:** Input Video + Step 1 Caption + Category Definition.
* **Task:**
    1. **Define a Target Event:** Write one short, atomic, directly observable statement of the interaction or consequence to be judged. Do not include negation, the answer, a category name, rationale, physical-validity language, or task-related meta-language.
    2. **Select a Typical Scenario:** For IV-A, select exactly `Trajectory Miss` or `Abrupt Halt`. For IV-B, select exactly `Insufficient Impact` or `Threshold Failure`.
    3. **Construct the Counterpart:** Describe a physically plausible paired video in which the same Target Event has the opposite occurrence status. If the original judgement is `Occurred`, the counterpart must be `Did Not Occur`; if the original judgement is `Did Not Occur`, the counterpart must be `Occurred`.
    4. **Store the Counterpart Scenario:** Write the counterpart description in `target_fallacy_scenario` while preserving its physical plausibility.

**Step 3: Generation Method Decision**
* **Basis:** Input Video + Step 2 Counterpart Scenario.
* **Principle for `Manual_CV_Edit`:** Select this when the counterpart can be achieved by editing, masking, freezing, compositing, cutting, temporal rearrangement, or reuse of existing pixels.
* **Principle for `AI_Generation`:** Select this when the counterpart requires synthesized motion, trajectories, interactions, action magnitude, consequences, state changes, objects, or scene content.
* **Task:** Select exactly one generation method without adding a separate rationale field.

### Output Format
*Return a single valid JSON object. Do NOT use markdown code blocks (like ```json).*

The value of `generation_method` must be exactly either `Manual_CV_Edit` or `AI_Generation`. Replace `{selected_generation_method}` with the selected value.

For IV-A, return exactly this field structure:

{
  "visual_fact_caption": "String. Strictly objective description of the original video's visible actions and objects.",
  "visual_style_and_camera": "String. E.g., 'Low-res CCTV, static camera' or '4K, shaky handheld'.",
  "interaction_judgement": "String. Exactly 'Occurred' or 'Did Not Occur'.",
  "target_event": "String. A short, atomic, directly observable initiating interaction shared by both videos.",
  "selected_typical_scenario": "String. Exactly 'Trajectory Miss' or 'Abrupt Halt'.",
  "target_fallacy_scenario": "String. Direct visual description of the physically plausible counterpart in which the Target Event has the opposite occurrence status.",
  "generation_method": "{selected_generation_method}"
}

For IV-B, return exactly this field structure:

{
  "visual_fact_caption": "String. Strictly objective description of the original video's visible actions and objects.",
  "visual_style_and_camera": "String. E.g., 'Low-res CCTV, static camera' or '4K, shaky handheld'.",
  "consequence_judgement": "String. Exactly 'Occurred' or 'Did Not Occur'.",
  "target_event": "String. A short, atomic, directly observable consequence or state change shared by both videos.",
  "selected_typical_scenario": "String. Exactly 'Insufficient Impact' or 'Threshold Failure'.",
  "target_fallacy_scenario": "String. Direct visual description of the physically plausible counterpart in which the Target Event has the opposite occurrence status.",
  "generation_method": "{selected_generation_method}"
}
\end{Verbatim}

\subsubsection{Near-Miss}
\begin{Verbatim}
### Role & Core Function
You are an expert in **Computer Vision (CV) and Generative AI**, specializing in video semantic understanding and synthetic data creation.

**Core Guidelines:**
* **Strict Adherence:** You must carefully read and strictly execute the `Category Definition` and `Reasoning Pipeline` provided by the user.
* **Visual Evidence Only:** All analysis must be strictly based on the visible pixel content of the video. Do NOT hallucinate objects or actions not present.
* **No Subjectivity:** **In the captioning phase**, strictly describe *what* is happening. Do NOT infer *why* (physics laws), *intent*, or *emotions*.
* **Contextual Adaptation:** The examples in "Typical Scenarios" are illustrative references. You must select the scenario that best matches the actual subjects and actions in the Input Video, define the core target event, and construct a precise counterpart in which that same event has the opposite occurrence status.

### Input Context
**Context 1: Category Definition:**
* **Core Principle:**
Consistency between Statistical Priors and Visual Facts. These are physically plausible paired videos designed to test whether a model judges a specified interaction from visible evidence instead of completing a familiar event script from semantic expectation.

* **Specific Adversarial Type: Near-Miss Interaction**
* **Definition:** A potential interaction is approached, but a visible spatial gap or a halt before the relevant contact prevents the initiating interaction in the non-occurring video. The paired video closes the gap and establishes the same target interaction. The target may be contact, trajectory intersection, object transfer, or action completion; transfer and completion refer here to interactions whose completion depends on closing the gap and establishing the relevant contact, not to arbitrary unfinished actions or downstream consequences after contact.
* **Typical Scenarios:**
    1.  **Trajectory Miss:** An object or person moves toward a target, but passes beside it with a visible spatial gap, preventing the target contact. The paired video shows the corresponding trajectories meeting, the gap closing, and the target interaction occurring.
    2.  **Abrupt Halt:** An object or person approaches the target but stops before the relevant contact while a visible spatial gap remains. The paired video shows the approach continuing until the gap closes and the specified interaction is completed.

**Context 2: Input Video:** The video file to be analyzed.

### Reasoning Pipeline (Chain-of-Thought)
*Perform the following 3 steps in a single logical flow:*

**Step 1: Visual Fact Anchoring**
* **Basis:** The Input Video.
* **Task A (Caption):** Generate a strictly objective visual description. Accurately describe the factual phenomena occurring in the video. Focus ONLY on **Subjects**, **Actions**, and **Environment**.
    * *Constraint:* Prohibition on explaining "why it happens" or "what it implies."
* **Task B (Style & Camera):** Extract visual style and camera movement information into a single field.
    * *Keywords:* Quality (CCTV/4K/Blurry/Motion Blur), Lighting, Camera Dynamics (Static/Pan/Zoom/Shaky Handheld).
* **Task C (Interaction Judgement):** Based strictly on visual evidence, determine whether the primary initiating interaction in the original video actually occurs.
    * *Constraint:* You must conclude with exactly "Occurred" or "Did Not Occur". Judge the specified initiating interaction, which may be contact, trajectory intersection, object transfer, or action completion. A transfer or completed interaction must be established by the visible closing of the relevant gap and contact evidence, not inferred from an approach alone. If the initiating action or contact has already occurred and only a later consequence is absent, that is the IV-B Consequence Arrest distinction rather than a spatial miss or pre-contact halt.

**Step 2: Target Event and Counterpart Scenario Construction**
* **Basis:** Input Video + Step 1 Caption + Category Definition.
* **Task:** Review Step 1's description and the user-provided `Category Definition`, then construct one matched counterpart that remains physically plausible.
    1. **Define a Target Event** as a short, atomic, directly observable statement of the initiating interaction to be judged. It must not contain an answer, negation, category name, rationale, or task-related meta-language.
    2. **Select** the most appropriate Typical Scenario (e.g., "Trajectory Miss").
    3. **Output a Target Relative Scenario** in `target_fallacy_scenario`. If the original interaction judgement is `Occurred`, the counterpart must show the Target Event not occurring. If it is `Did Not Occur`, the counterpart must show the Target Event occurring. Describe the visible counterpart directly without introducing any physical violation.

**Step 3: Generation Method Decision**
* **Basis:** Input Video + Step 2 Target Relative Scenario.
* **Task:** Select `Manual_CV_Edit` when the counterpart can be created by editing, masking, freezing, compositing, cutting, or rearranging existing pixels. Select `AI_Generation` when it requires synthesizing new motion, trajectories, interactions, objects, or scene content.

### Output Format
*Return a single valid JSON object. Do NOT use markdown code blocks (like ```json).*

The value of `generation_method` must be exactly either `Manual_CV_Edit` or `AI_Generation`. Replace `{selected_generation_method}` with the selected value.

{
  "visual_fact_caption": "String. Strict objective description of actions and objects.",
  "visual_style_and_camera": "String. E.g., 'Low-res CCTV, static camera' or '4K, shaky handheld'.",
  "interaction_judgement": "String. Exactly 'Occurred' or 'Did Not Occur'.",
  "target_event": "String. A short, atomic, directly observable initiating interaction shared by both videos.",
  "selected_typical_scenario": "String. The exact title of the chosen scenario, e.g., 'Trajectory Miss'.",
  "target_fallacy_scenario": "String. Direct visual description of the physically plausible counterpart in which the Target Event has the opposite occurrence status.",
  "generation_method": "{selected_generation_method}"
}
\end{Verbatim}

\subsubsection{Consequence Arrest}
\VerbatimInput[fontsize=\scriptsize,breaklines=true,breakanywhere=true]{prompts/prompt_IVB.md}

\subsection{Video Generation Prompt Construction}
\label{appendix:prompt_generation}

\subsubsection{Anti-physics Counterpart Generation}
\begin{Verbatim}
### Dynamic Inputs
* **Visual Frames:** [Input: 16 extracted frames from the original video]
* **Camera & Style:** [Input: Extracted visual style and camera movement]
* **Original Caption:** [Input: Strict objective description of the original action]
* **Target Fallacy Scenario:** [Input: Detailed description of the target physics violation]

### Core Task & Directives
Based on the `Visual Frames`, `Camera & Style`, and `Original Caption` of the source video, you must write a highly precise English prompt for a video generation model to guide it in generating the given `Target Fallacy Scenario`, creating a modified new version of the video.

* **Establish Baseline & Execute Replacement:** The prompt must use the `Original Caption` as the realistic baseline for the environment and subjects, but you must **completely replace** the logical, natural outcome of the original video with the physically invalid events described in the `Target Fallacy Scenario`.
* **Absolute Forcing Principle:** The prompt must force the video model to prioritize executing the `Target Fallacy Scenario`. You are **strictly prohibited** from attempting to "correct" or rationalize these anti-physical phenomena. You must command these impossible events directly as objective visual facts.

### Core Prompting Rules
* **Structural Front-loading:** Always place the camera movement, visual style, or perspective (e.g., "High quality, Low angle shot") at the very beginning of the prompt to set the visual tone.
* **Strong Explicit Negation & Contrast:** You MUST use strong negative words (e.g., capitalized "NOT", "without") or extreme contrasting modifiers (e.g., "tiny pebble" vs. "towering geyser") to definitively break common sense and semantic priors.
* **Temporal Slicing Expression:** For sudden state changes or interventions, use clear temporal anchors to define the sequence of events (e.g., "Upon impact", "Phase 1...", "Trigger:", "Phase 2...").
* **Summary Qualitative Label:** Conclude the entire prompt with an extremely brief declarative sentence that definitively labels the physical anomaly (e.g., "The material behaves like rubber.").

### Few-Shot Examples
[Example 1] High-speed slow-motion footage with detailed deformation texture. A pink vase falls and hits the rusty floor. Upon impact, the rigid vase exhibits soft-body behavior: it drastically deforms, squashes, and compresses, then rebounds elastically and recovers its shape. The material behaves like rubber.

[Example 2] High quality, realistic footage. Dynamic tracking shot matching the original camera movement. Phase 1: The original dog is walking/running on the grass. It maintains its original posture and movement rhythm. Trigger: Mid-stride, a surreal transformation occurs. Phase 2: The animal smoothly morphs into a ginger tabby cat. The cat occupies the exact same position and path as the dog. The cat continues the walking motion seamlessly. Environment Constraint: The grass texture, lighting, and background scenery remain identical to the original footage. Only the animal species changes.

[Example 3] High quality, realistic footage starting from the reference image. The metal knife blade presses down firmly into an ordinary soft mandarin. Upon cutting pressure, the fruit does NOT split along the blade like soft, juicy flesh. Instead, it behaves like a brittle ceramic object, instantly fracturing into more than six small, separate pieces and irregular peel fragments that scatter across the textured glass cutting board. Juice sprays slightly. The knife continues downwards to the board. The ordinary fruit exhibits an anomalous brittle material response.

[Example 4] Low angle shot. A tiny pebble drops vertically into the center of the calm lake. The moment it touches the surface, the water erupts violently. A massive, towering geyser of water shoots fifty meters high into the air immediately upon contact. The impact site generates a giant explosion of white foam and heavy waves radiating outwards. The scene demonstrates extreme kinetic amplification.

[Example 5] Low angle shot. A massive grey boulder falls from the sky and hits the water surface. The water surface remains completely undisturbed and flat.

### Output Constraint
* Return a single valid JSON object. Do NOT use markdown code blocks (like ```json).
* The JSON must contain exactly one key: "generated_video_prompt".

{
  "generated_video_prompt": "String. The final constructed English prompt."
}
\end{Verbatim}

\subsubsection{Counter-intuitive Counterpart Generation}
\VerbatimInput[fontsize=\scriptsize,breaklines=true,breakanywhere=true]{prompts/prompt_generate_IV.md}

\subsection{O--A--V Annotation Synthesis}
\label{appendix:prompt_annotation}

\subsubsection{Generated Anti-physics Counterparts}
\VerbatimInput[fontsize=\scriptsize,breaklines=true,breakanywhere=true]{prompts/label_prompt.md}

\subsubsection{Coherence Violation}
\VerbatimInput[fontsize=\scriptsize,breaklines=true,breakanywhere=true]{prompts/label_prompt_IA.md}

\subsubsection{Causal Reversal}
\VerbatimInput[fontsize=\scriptsize,breaklines=true,breakanywhere=true]{prompts/label_prompt_IB.md}

\subsubsection{Continuous-Reversal Dynamic Violations}
\VerbatimInput[fontsize=\scriptsize,breaklines=true,breakanywhere=true]{prompts/label_prompt_rev.md}

\subsubsection{Counter-intuitive Events}
\VerbatimInput[fontsize=\scriptsize,breaklines=true,breakanywhere=true]{prompts/label_prompt_IV.md}

\subsection{PriorPair Task Prompts}
\label{appendix:prompt_tasks}
\VerbatimInput[fontsize=\scriptsize,breaklines=true,breakanywhere=true]{prompts/task_prompts.md}

\subsection{RAS Evaluation Prompts}
\label{appendix:prompt_ras}
\VerbatimInput[fontsize=\scriptsize,breaklines=true,breakanywhere=true]{prompts/ras_evaluation_prompts.md}

\subsection{SPD Diagnostic Prompts}
\label{appendix:prompt_spd}
\VerbatimInput[fontsize=\scriptsize,breaklines=true,breakanywhere=true]{prompts/semantic_prior_diagnosis_prompts.md}

\end{document}